\pdfoutput=1

\documentclass[11pt]{article}

\usepackage{acl}

\usepackage{times}
\usepackage{latexsym}

\usepackage[T1]{fontenc}

\usepackage[utf8]{inputenc}

\usepackage{microtype}

\usepackage{inconsolata}

\usepackage{subcaption}
\usepackage{adjustbox}
\usepackage{graphicx}
\usepackage{booktabs}
\usepackage{multirow}
\usepackage{amsmath}
\usepackage{bm}
\usepackage{tabularx}
\usepackage{array}
\usepackage{amssymb}
\usepackage{makecell}

\newcommand*{\belowrulesepcolor}[1]{%
  \noalign{%
    \kern-\belowrulesep
    \begingroup
      \color{#1}%
      \hrule height\belowrulesep
    \endgroup
  }%
}

\newcommand*{\aboverulesepcolor}[1]{%
  \noalign{%
    \begingroup
      \color{#1}%
      \hrule height\aboverulesep
    \endgroup
    \kern-\aboverulesep
  }%
}

\definecolor{my-blue}{HTML}{eaeefb} 
\definecolor{my-gray}{HTML}{EEEEEE}
\definecolor{my-green}{HTML}{43962a}
\newcommand{\methodname}{LLM-to-SLM}

\definecolor{linksblue}{rgb}{0.196, 0.325, 0.863}
\hypersetup{colorlinks, linkcolor=linksblue, citecolor=linksblue}

\title{Think Big, Generate Quick: \\ \methodname{} for Fast Autoregressive Decoding}

\author{
Benjamin Bergner$^{1,2,\dagger,*}$ \quad Andrii Skliar$^{2,*}$ \\
{\bf Amelie Royer$^{2,\ddagger}$} \quad {\bf Tijmen Blankevoort$^{2,\ddagger}$} \quad {\bf Yuki Asano$^{2,3}$} \quad {\bf Babak Ehteshami Bejnordi$^{2}$} \\
$^{1}$Hasso Plattner Institute for Digital Engineering, University of Potsdam \\
$^{2}$Qualcomm AI Research, Amsterdam $^{3}$QUVA Lab, University of Amsterdam\\
\texttt{benjamin.bergner@hpi.de} \quad \texttt{y.m.asano@uva.nl} \\
\texttt{\{askliar, aroyer, tijmen, behtesha\}@qti.qualcomm.com} \\
}
\begin{document}
\maketitle

{\let\thefootnote\relax\footnotetext{$*$Equal contribution $^\dagger$Work done as an intern at Qualcomm AI Research $^\ddagger$Work was completed while employed at Qualcomm AI Research; Qualcomm AI Research is an initiative of Qualcomm Technologies, Inc. Correspondence to: Andrii Skliar, <askliar@qti.qualcomm.com>.}}
{\let\thefootnote\relax\footnotetext{\textit{Work presented at the ES-FoMo II Workshop at ICML 2024},
Vienna, Austria.
Copyright 2024 by the author(s).}}

\begin{abstract}
Large language models (LLMs) have become ubiquitous in practice and are widely used for generation tasks such as translation, summarization and instruction following.
However, their enormous size and reliance on autoregressive decoding increase deployment costs and complicate their use in latency-critical applications. 
In this work, we propose a hybrid approach that combines language models of different sizes to increase the efficiency of autoregressive decoding while maintaining high performance.
Our method utilizes a pretrained frozen LLM that encodes all prompt tokens once in parallel, and uses the resulting representations to condition and guide a small language model (SLM), which then generates the response more efficiently.
We investigate the combination of encoder-decoder LLMs with both encoder-decoder and decoder-only SLMs from different model families and only require fine-tuning of the SLM.
Experiments with various benchmarks show substantial speedups of up to $4\times$, with minor performance penalties of $1-2\%$ for translation and summarization tasks compared to the LLM.

\end{abstract}


\section{Introduction}

The recent widespread adoption of large language models (LLMs) has enabled a variety of applications in the field of natural language generation (NLG), from machine translation~\cite{wu2016google} and code completion~\cite{chen2021evaluating} to general-purpose chatbots
~\cite{openai2023gpt4}.
Their performance is a function of compute, dataset size and parameter count~\cite{kaplan2020scaling, hoffmann2022training}, with emerging abilities becoming apparent only at large scales~\cite{thoppilan2022lamda, chowdhery2022palm, wei2022emergent}.
These findings have led to the increased popularity of large models, both in decoder-only~\cite{scao2022bloom, zhang2022opt, touvron2023llama} and encoder-decoder networks~\cite{chung2022scaling}.

As this race to scale intensifies, LLMs are becoming challenging to deploy, especially in light of compute limitations and latency requirements on edge devices, which ultimately translate into higher costs for providers and end users alike~\cite{chen2023frugalgpt}.
More specifically, LLMs in NLG operate in two phases:
(1) First, encoding the user prompt (e.g., \texttt{Translate into German: I love you}), followed by (2) decoding of the response (\texttt{Ich liebe dich}). 
In many cases, such as translation or summarization,  the prompt is known in advance and can thus be processed efficiently in parallel. However, the response is usually generated in an autoregressive, sequential manner~\cite{radford2018improving, zarriess2021decoding}: 
The LLM must be called for each token to be generated, requiring all its weight matrices and the KV cache to be loaded.
As a result, decoding becomes bound to the memory bandwidth of the accelerator, which eventually leads to high inference latency as the length of the response grows~\cite{pope2023efficiently}.

\begin{figure}[tbp]
\includegraphics[width=.99\linewidth]{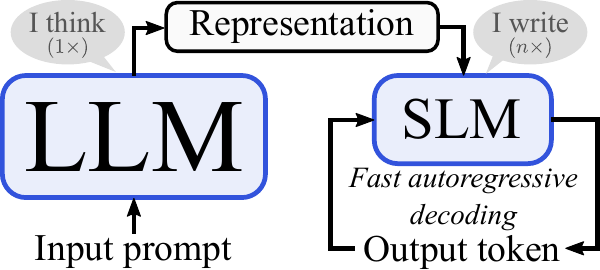}
\caption{\methodname: A large language model (LLM) computes a high-quality representation of the prompt to condition a small language model (SLM), which then efficiently decodes the response while maintaining high performance close to the LLM.}
\label{fig:llm2slm_highlevel}
\end{figure}

Research aimed at reducing the overall inference cost of LLMs has garnered significant interest over the past few years.
While traditional techniques, e.g. in model compression and parallel decoding, are still active areas of research, recent developments show a pivot towards hybrid approaches that combine models of different sizes for fast decoding, such as speculative decoding~\cite{leviathan2023fast, chen2023accelerating} or LLM cascades~\cite{chen2023frugalgpt}.
Despite these promising developments, exploiting the discrepancy between the fast prompt encoding phase and the slow response generation remains underexplored. 
%
%
%
%
%

Intuitively, having a detailed understanding of the prompt is critical for the planning and delivery of an appropriate response.
In contrast, autoregressive decoding aims at predicting the next token, which is comparatively low-level and can in certain settings be accomplished well even by SLMs~\cite{eldan2023tinystories}.
Following this insight, we propose to reduce the cost of autoregressive decoding with a hybrid model in which these complementary tasks are distributed over two unequally sized networks (\hyperref[fig:llm2slm_highlevel]{Figure~\ref{fig:llm2slm_highlevel}}).
Specifically, we perform a single forward pass with an LLM to compute a high-quality representation of the prompt, which is used to condition a more efficient SLM that then performs autoregressive generation.
Since the prompt can be encoded in parallel, computing an LLM representation to guide an SLM results in only a minor increase in overall runtime compared to the SLM alone, in particular for tasks such as machine translation that may require generating longer sequences. We show that this minor increase in runtime compared to the SLM allows for a substantial increase in predictive performance.
Overall, we make the following contributions:
\begin{itemize}
    \item We present \methodname, a simple approach for fast autoregressive decoding where an LLM conditions an SLM. We mix encoder-decoder LLMs with both encoder-decoder and decoder-only SLMs and only require fine-tuning of the SLM.
    \item We empirically evaluate the efficacy of different \methodname{} variants in various task domains, training and evaluation regimes: traditional fine-tuning for both machine translation and summarization, as well as instruction-tuning evaluated in the zero-shot setting on multiple held-out tasks.
    \item Our method accelerates pretrained LLMs while maintaining high performance. In translation and summarization, LLM-to-SLM achieves a speedup of $4.2\times$ and $3.0\times$, respectively, with a marginal drop of $1-2\%$ in predictive performance.
    
\end{itemize}

\section{Related Work}

The duality of model performance and cost has sparked a lot of research interest in LLM efficiency, which is approached from various angles.

\paragraph{Model compression.} 
%
A common approach to accelerate LLMs is to create a simpler compressed version using pruning~\cite{frantar2023sparsegpt, ma2023llm, sun2023simple} or quantization ~\cite{dettmers2022llm, yao2022zeroquant, xiao2023smoothquant}. 
A third pillar of model compression is knowledge distillation where a small model learns from outputs of larger models~\cite{hinton2015distilling}.
For the language domain, chain-of-thought prompting~\cite{wei2022chain} can be applied to generate samples from an LLM teacher, that can subsequently be used by an SLM student as training signal~\cite{ho2022large, magister2022teaching, li2023symbolic, shridhar2023distilling, hsieh2023distilling}.
\methodname{} can be regarded as a model compression technique, where the decoder responsible for generation is replaced by a smaller, more efficient model, while retaining the LLM for prompt encoding.

\paragraph{Parallel decoding.}
Instead of predicting one token at a time, parallel decoding aims at generating multiple tokens at once~\cite{gu2017non,wang2019non, sun2019fast, wei2019imitation}.
While empirical speedups are possible, this approach raises new challenges such as inconsistencies in the output and the need to estimate the length of the response, as language is inherently conditional.
In practice, it may require multiple iterations~\cite{ghazvininejad2019mask, lee2018deterministic, stern2018blockwise, santilli2023accelerating, fu2023lookahead}, or a combination of parallel and sequential decoding~\cite{ning2023skeleton}.
Our method sticks to the more commonly used autoregressive decoding but uses a small model to decrease inference time.

\paragraph{Conditional approaches.}
In comparison to the language domain, where conditioning of multiple networks is still relatively rare, it is used to a greater extent in multimodal learning.
For example, various works have conditioned a language model on vision features~\cite{driess2023palm, chen2022pali, zhou2022conditional, liu2023visual}.
Interestingly, ~\citet{driess2023palm} and~\citet{liu2023visual} study freezing parts of their models, which is related to how we freeze the LLM during fine-tuning.
In vision, there is also a growing trend of conditioning small transformer decoders on larger encoders~\cite{bergner2022iterative, he2022masked, jaegle2021perceiver}.
Instead, our method connects a small network to a large network for NLG.

%
\paragraph{Learned prompts.}
Our method is related to parameter-efficient fine-tuning (PEFT) techniques that incorporate trainable prompts, which can be both continuous~\cite{lester2021power, li2021prefix, liu2023gpt} or discrete~\cite{deng2022rlprompt, prasad2022grips, zhang2022tempera}.
In particular, ~\citet{lester2021power} prepend soft prompts to the input and freeze all other parameters.
This is similar to the way we fuse the LLM representation into the SLM.
We experiment with both trainable and fixed SLMs, with only a projector being trained in the latter case.
In contrast to prompting methods, our main focus is on reducing inference cost.
\paragraph{SLMs.}
The most straightforward route to efficiency is smaller models.~\citet{schick2020s} showed that SLMs can do few-shot learning and outperform GPT3 in the SuperGLUE benchmark.
Another direction is to train language models on limited vocabulary~\cite{warstadt2023call, huebner2021babyberta}.
Notably, TinyStories~\cite{eldan2023tinystories} learns various SLMs on LLM-generated stories using vocabulary that a 4-year-old child can normally understand and demonstrates coherent English-generated text.
While these works have much future potential, we are investigating a more practical hybrid model with wide applicability in NLG tasks that combines high performance of LLMs with efficiency of SLMs.
\paragraph{Hybrids.}
Most related to our work are hybrid models that employ both LLM and SLM, where the latter performs the bulk of computation.
~\citet{chen2023frugalgpt} propose a language model cascade, where cheaper models are invoked first.
~\citet{jiang2023recyclegpt} recycle the representation of the LLM by passing it to a smaller model that predicts the subsequent token more efficiently. 
Speculative decoding (SD) methods repeatedly call the SLM to generate a draft that is then validated in parallel by the LLM~\cite{chen2023accelerating, leviathan2023fast, kim2023big}.
Medusa~\cite{cai2023medusa} is related but attaches heads on top of the LLM to predict multiple tokens in parallel. 
During generation, the LLM in SD is invoked several times, whereby the frequency of LLM calls depends on the performance of the SLM.
In our method, the LLM is only called once and the SLM is conditioned on its representation.

Note that our method is orthogonal to most of these related works. For example, one may combine our method with quantization, pruning or other hybrid approaches. 
In \hyperref[ss:spec_dec]{Section~\ref{ss:spec_dec}}, we demonstrate that our approach can be combined together with speculative decoding. 

%







\section{\methodname}


\begin{figure}[tbp]
\includegraphics[width=1.\linewidth]{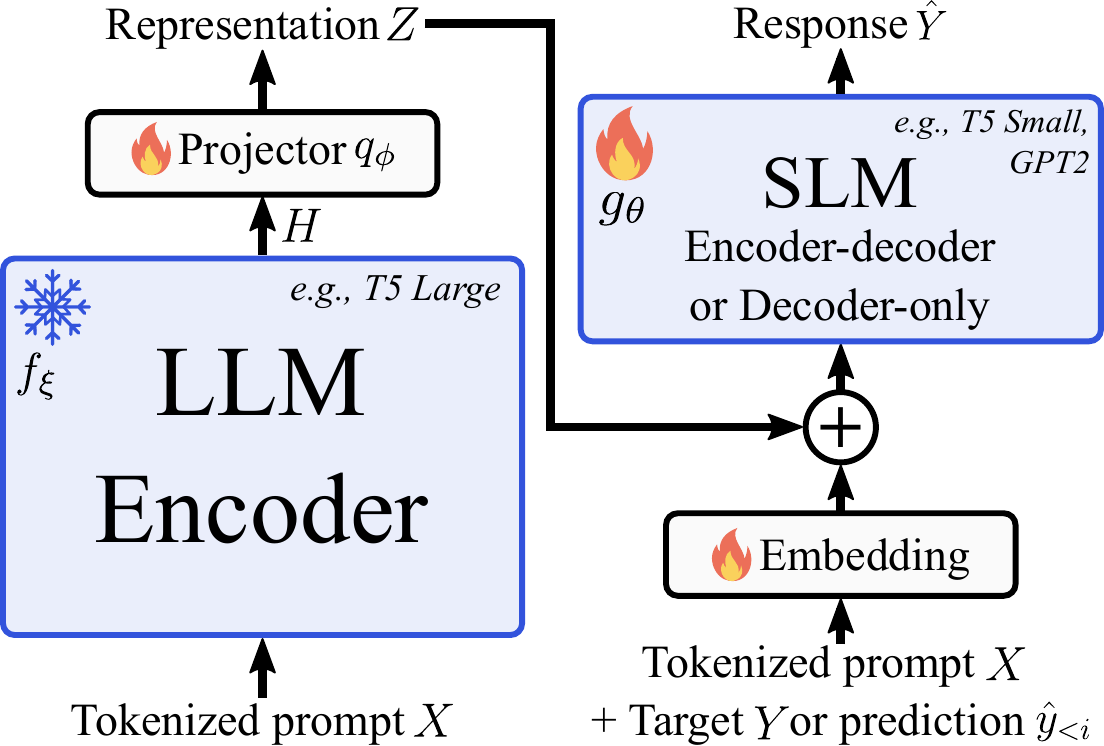}
\caption{Architecture details. A frozen LLM encoder integrates projected representations into either a trainable encoder-decoder or a decoder-only SLM.}
\label{fig:llm2slm_lowlevel}
\end{figure}

It is well established that model capacity and overparametrization play a crucial factor in model performance~\cite{kaplan2020scaling, hoffmann2022training}.
Following this insight, the core idea of \methodname{} is to compensate the low parameter count of an SLM by conditioning its next token prediction on a high-quality representation of the prompt given by an LLM. 

\hyperref[fig:llm2slm_lowlevel]{Figure~\ref{fig:llm2slm_lowlevel}} presents an overview of \methodname:  First, the  \textbf{LLM} encoder $f_{\xi}$ computes a high-quality representation of the prompt.
The \textbf{projector} $q_{\phi}$ then adapts and projects this representation to the SLM embedding space.
Finally, the \textbf{SLM} $g_{\theta}$ takes the projected representation and the prompt as input and generates the output tokens in an autoregressive manner.
Crucially, the parameter count of the SLM is significantly smaller than the LLM  ($8-55\times$ in our experiments), leading to faster generation as only the SLM performs autoregressive decoding.
In the remainder of this section, we further describe the individual modules of \methodname{} and explain how the representations of the LLM are injected into both encoder-decoder and decoder-only SLMs.



\subsection{Fast autoregressive decoding}

Given a prompt $X=\left[x_1,\dots,x_m\right]$ and an encoder-decoder LLM, 
autoregressive decoding models the output $Y=\left[y_1,\dots,y_n\right]$ in a causal manner:
\begin{equation}
    p(Y\vert X)=\prod_{i=1}^{n} g_\xi(y_i\vert y_{<i}, f_\xi(x_{1:m})),
\end{equation}
where $f_\xi$ and $g_\xi$ refer to the LLM encoder and decoder, respectively.
%
%
%
Generating the complete sequence $Y$ thus requires $n$ very costly forward  passes to the LLM decoder $g_\xi$. 
Furthermore, these calls can not be parallelized as we need to first sample the token $y_i$ to estimate the probability distribution over the $i + 1$-th token.
Instead, we propose to delegate the costly autoregressive decoding calls to a smaller language model, while preserving the encoder capacity:
\begin{equation}
    p(Y\vert X)=\prod_{i=1}^{n} g_{\theta}(y_i\vert y_{<i}, x_{1:m}, q_{\phi}( f_{\xi}(x_{1:m}))).
    \label{eq:llm2slm}
\end{equation}
The LLM is now only called once to provide a high quality encoding of the input prompt  to the SLM.
Therefore, as the number of autoregressive steps $n$ increases, the runtime of our method converges towards the original runtime of the SLM. 

\subsection{Architecture}
\subsubsection{LLM encoder}
\label{sss:llm_encoder}
The LLM encoder $f_{\xi}:X\mapsto H$ takes a prompt $X$ of length $m$ as input and computes a high-quality representation $H\in\mathbb{R}^{m\times d_l}$ of it.
In training LLMs, the requirement for substantial computational resources is often a limiting factor.
In our approach, we focus on a more resource-efficient training regime closer to that of the SLM, by freezing the parameters of the LLM during fine-tuning.
In this way, we can train \methodname{} on a small budget by pre-computing prompt representations.
To ensure good representation quality, we leverage large pretrained encoder-decoder models, but omit the decoder. The last layer representation of the encoder in an encoder-decoder architecture serves as a straightforward prompt encoding point for an SLM. In a decoder-only model, in contrast, determining the exact intermediate layer(s) that contribute to encoding of the prompt is more challenging (see \hyperref[sec:appendix_decoder_only]{Appendix \ref{sec:appendix_decoder_only}}), as the last layer representation is mainly useful for the low-level task of predicting the next token~\cite{chen2020generative}. 
\subsubsection{Projector}
The projector $q_{\phi}:H\mapsto Z$ has the task of aligning representations between LLM and SLM.
It takes high-dimensional features $H\in\mathbb{R}^{m\times d_l}$ of the LLM as input and computes a lower-dimensional representation $Z\in\mathbb{R}^{m\times d_s}$ which can be fused directly with embeddings of the SLM. 
Although there are many ways to design such a projector, we found that using a small MLP: \texttt{Linear($d_l$, $d_s$)} $\to$ \texttt{ReLU} $\to$ \texttt{Linear($d_s$, $d_s$)} trained from scratch is simple, efficient and performs well.

\subsubsection{SLM}
\label{sss:slm}
The SLM $g_{\theta}:(X,Z)\mapsto\hat{Y}$ maps the tokenized input $X$ and the projected representation $Z\in\mathbb{R}^{m\times d_s}$ to the response $\hat{Y}$ of length $n$.
During training, we also append the target sequence $Y$ to the input for next token prediction via teacher forcing~\cite{williams1989learning}.
Equivalently, predicted tokens $\hat{y}_{<i}$ are appended during inference.
We employ pretrained networks as SLMs, but fine-tune them as they have not been previously trained to process high-capacity encodings.
Furthermore, we perform experiments in which we learn either from the ground truth signal provided by a dataset or from sequences generated by the LLM.

\paragraph{Feature integration.}
A critical design decision in our framework is the way in which the SLM is conditioned.
We intend to integrate LLM representations into the SLM at an early stage, as this allows us to treat the internal structure of the SLM as a black box and use both encoder-decoder and decoder-only SLMs in a simple and unified way.
Initially, the tokenized input ($X; Y \text{or } \hat{y}_{<i}$) is embedded using a trainable embedding matrix.
A straightforward way to fuse features is then to \textit{replace} the SLM embedding of the prompt with the projected LLM representation.
This is similar to soft prompt tuning~\cite{lester2021power}, with the difference that our prompts are conditioned on LLM features.
An alternative strategy used in the main experiments is \textit{adding} the projected LLM representation $Z$ onto the SLM prompt embedding $E_X$, such that $Z+E_X$ is the input to the SLM.
In this way, we preserve the semantics between SLM embeddings of prompt and target/predicted sequence and leverage the LLM representation to modulate the prompt embedding via addition (see \hyperref[ss:ablation]{Section~\ref{ss:ablation}} for a comparison of the two approaches).

\paragraph{Aligning sequence lengths.}
When fusing the LLM representation with the SLM embedding of the prompt via addition, sequence lengths must be aligned, which is not guaranteed when combining models from different families that may use different tokenizers and vocabularies.
To this end, we propose to reuse the LLM tokenizer and embedding matrix to align the sequence lengths such that only a single tokenizer needs to be maintained.
In this case, we employ two new linear layers: (1) an embedding projection layer that maps these LLM embeddings from dimension $d_l$ to the SLM embedding space of dimension $d_s$ and (2) a new head layer that replaces the original SLM head and maps to the vocabulary used by the LLM.
Finally, note that there are various other options for fusing the features of different models, e.g. cross-attention, or FiLM layers~\cite{perez2018film}.
However, we opted for adding, as it is simple and requires minimal changes to the SLM.
\section{Experiments}
In this section, we intend to answer the following question: What is the comparative performance and runtime of our proposed \methodname{} method in relation to LLM and SLM alone? 
To this end, we first empirically evaluate the efficacy of \methodname{} on three tasks: machine translation, summarization and instruction tuning. We then report the computational efficiency of our proposed method in \hyperref[ss:runtime]{Section~\ref{ss:runtime}}. Finally, in \hyperref[ss:ablation]{Section~\ref{ss:ablation}} we present a comprehensive set of ablation studies, investigating the performance of \methodname{} under varying SLM capacities, its orthogonality to speculative decoding, and how it compares to PEFT methods.

\paragraph{Setup.}
The networks used in our experiments are listed in \hyperref[table:models]{Table~\ref{table:models}}.
We employ various pretrained models and architectures and denote combinations as LLM$\,\to\,$SLM.
We make use of T5 encoders as LLMs, and employ T5 encoder-decoder and GPT2 decoder-only models as SLMs: 
In this setting, the LLMs have $8-55\times$ more parameters than the SLMs.
Since the LLM remains frozen in our method, we rely on well-performing LLMs pre-trained for the respective tasks, such that the specific model selection may vary in the investigated benchmarks.
For generation, we use beam search (beam width of $4$, length penalty of $0.6$) for translation and summarization, and nucleus sampling for instruction tuning.
We report task-specific performance metrics: SacreBLEU~\cite{post-2018-call} for translation, ROUGE~\cite{lin2004rouge} for summarization and GPT4 as a judge for instruction tuning, using 
the same generation settings as~\cite{zheng2023judging}.
Furthermore, we report runtimes per single generated token (in milliseconds). These are calculated from generating a total of $100$ tokens with a prompt length of also $100$ tokens, either on an NVIDIA V100 (translation, summarization) or NVIDIA A100 (instruction tuning) GPU. 

\begin{table}
\centering
\noindent\adjustbox{max width=\linewidth}{
\begin{tabular}{lll}
\toprule
& \textbf{Model} & \textbf{Params}\\
\midrule
\multirow{6}{*}{{\rotatebox[origin=c]{90}{Enc-Dec}}}
& T5 Small$^\dagger$~\cite{raffel2020exploring} & 44M (19M/25M) \\
& T5 1.1 Small$^\dagger$~\cite{raffel2020exploring} & 44M (19M/25M) \\
& T5 Large$^*$~\cite{chung2022scaling} & 737M (302M, 402M) \\
& Flan T5 Base$^\dagger$~\cite{chung2022scaling} & 198M (85/113M) \\
& Flan Alpaca Base~\cite{Chia_Hong_Poria_2023} & 198M (85/113M) \\
& Flan T5 XXL~\cite{chung2022scaling} & 10.9B (4.6/6.2B) \\
& Flan Alpaca XXL$^*$~\cite{Chia_Hong_Poria_2023} & 10.9B (4.6/6.2B) \\
\midrule
\multirow{4}{*}{{\rotatebox[origin=c]{90}{Dec-only}}}
& GPT2$^\dagger$~\cite{radford2019language} & 86M\\
& GPT2 [1,2,4]-Layers$^\dagger$ & [8M, 15M, 29M]\\
& GPT2 XL~\cite{radford2019language} & 1.5B\\
& LLaMA 13B~\cite{touvron2023llama} & 12.7B\\
\bottomrule
\end{tabular}
}
\caption{Model variants used in the experiments. Sizes are rounded, excluding embedding and head parameters. Encoder/decoder sizes are shown in parentheses. Symbols $*$ and $\dagger$ denote models that we use in our method as LLMs and SLMs respectively.}

\label{table:models}
\end{table}

\paragraph{Training.}
All models are trained with an  effective batch size of 128, cross-entropy loss, AdamW optimizer~\cite{loshchilov2017decoupled} with weight decay of $0.1$, learning rate of $0.001$ with linear warmup~\cite{goyal2017accurate} for $10\%$ of the total number of iterations, followed by cosine learning rate decay to $0$~\cite{loshchilov2016sgdr}.
We rely on Huggingface's transformers~\cite{wolf2019huggingface} for training and generation.
Further training details are listed in \hyperref[sec:appendix_model_training]{Appendix \ref{sec:appendix_model_training}}.











\subsection{Machine translation}

We report results for the translation task from English to German, French, and Romanian.
We utilize WMT14~\cite{bojar2014findings} for En-Fr/De, and WMT16 for En-Ro~\cite{bojar2016findings}.
T5 Large serves as LLM encoder, and T5 Small, T5 1.1 Small, and GPT2 as SLMs.
T5 Small and GPT2 are $16\times$ and $8\times$ smaller than T5 Large. 
Both T5 Large/Small are pretrained for translation, while T5 1.1 Small was only trained on the C4 dataset.
The following task description precedes the prompt: "\texttt{translate English to *target-language*:}".
All models are trained for $50k$ iterations, except for T5 Large which comes pretrained for translation and is applied in a zero-shot manner. 
We use our LLM to generate training labels. 
We report BLEU scores evaluated on the test splits in \hyperref[tab:translation]{Table~\ref{tab:translation}}. 

\begin{table}[t]
\centering
\noindent\adjustbox{max width=\linewidth}{
\begin{tabular}{llllll}
\toprule
\textbf{Model} & \textbf{En-Fr} & \textbf{En-De} & \textbf{En-Ro} & \textbf{Avg.} & \textbf{Time}\\
\midrule 
T5 Large (zero-shot) & 39.53 & 29.10 & 27.19 & 31.94 & 61.5 \\
\midrule
\midrule
T5 Small & 37.16 & 26.47 & 26.15 & 29.93 & 14.2 \\
T5 1.1 Small & 34.85 & 24.55 & 25.25 & 28.22 & 18.4 \\
GPT2 & 35.67 & 25.53 & 24.70 & 28.63 & 19.7 \\
\midrule
\belowrulesepcolor{my-blue}
\rowcolor{my-blue}
T5 Large$\,\to\,$T5 Small & 39.22 & 28.36 & 27.04 & 31.54 & 14.8 \\
\rowcolor{my-blue}
T5 Large$\,\to\,$T5 1.1 Small & 38.21 & 27.42 & 26.54 & 30.72 & 19.1 \\
\rowcolor{my-blue}
T5 Large$\,\to\,$GPT2 & 39.01 & 28.27 & 25.86 & 31.05 & 20.7 \\
\aboverulesepcolor{my-blue}
\bottomrule
\end{tabular}
}
\caption{BLEU scores for machine translation. LLM-to-SLM models approach the performance of the LLM.} 
\label{tab:translation}
\end{table}
T5 Large shows an average score of $31.94$, while T5 Small performs $2$ BLEU points worse. SLMs that have not been previously pretrained for translation (T5 1.1 Small and GPT2) perform more than $3$ BLEU points worse than the LLM.
In comparison, our \methodname{} variants reduce this gap to less than 1 point: 
For example, T5 Large$\,\to\,$T5 Small achieves on average a BLEU score of $31.54$ across languages. 
Furthermore, gains over the SLM baseline are more pronounced when using pretrained networks that were not previously trained on translation, e.g., T5 Large$\,\to\,$T5 1.1 Small scores on average $2.5$ points better than T5 1.1 Small (8.8\% increase).
Finally, despite having different model families, we also observe significant gains in T5 Large$\,\to\,$GPT2 (8.4\% increase), and the task performance improvements come at a marginal trade-off in runtime. 

\begin{figure*}[htb]
  \centering
  \begin{subfigure}[b]{0.33\textwidth}
    \includegraphics[width=\textwidth]{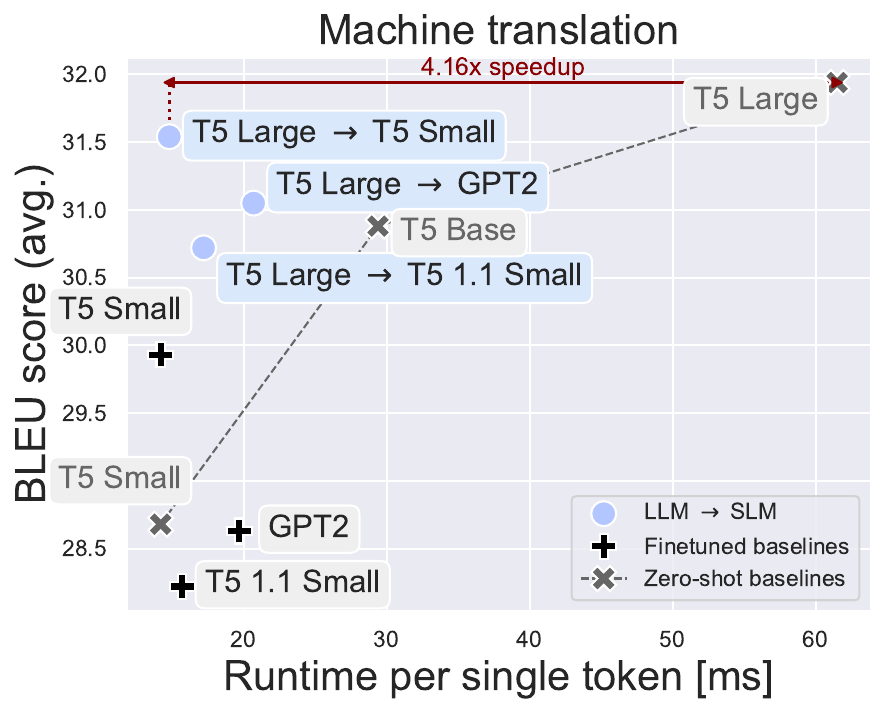}
    \label{fig:1}
  \end{subfigure}
  \hfill
  \begin{subfigure}[b]{0.32\textwidth}
    \includegraphics[width=\textwidth]{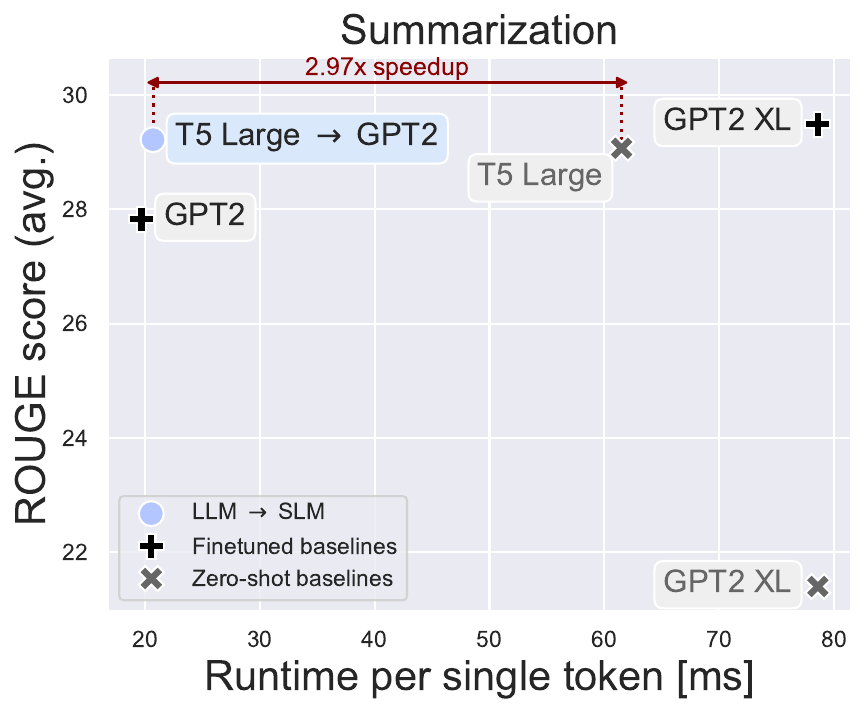}
    \label{fig:2}
  \end{subfigure}
  \hfill
  \begin{subfigure}[b]{0.33\textwidth}
    \includegraphics[width=\textwidth]{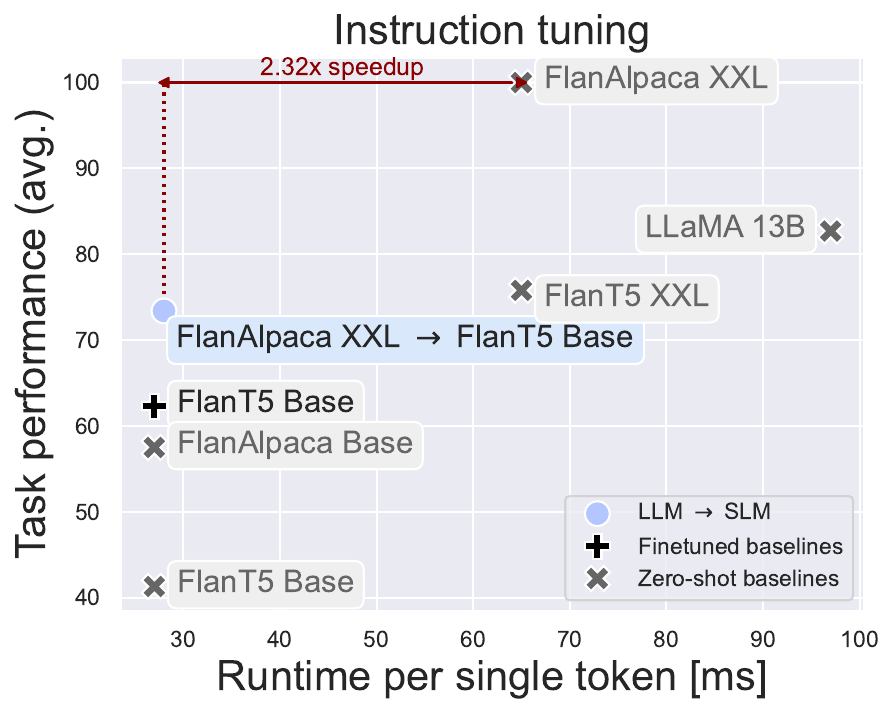}
    \label{fig:3}
  \end{subfigure}
  \vspace{-1\baselineskip}
  \caption{Performance-runtime trade-off curves for various models across different tasks.}
  \label{fig:three-figures}
\end{figure*}



\subsection{Summarization}

We further assess the performance of \methodname{} when combining models from different families for the task of  summarization on CNN/Daily Mail~\cite{hermann2015teaching}.
We again use T5 Large as LLM and GPT2 as SLM.
Similar to translation, T5 Large comes pretrained for summarization.
Following~\citet{raffel2020exploring}, we build the input prompt by prefixing the input text with "\texttt{summarize:}".
We fine-tune all models for $25k$ iterations on the training set and evaluate on the test split. 
In contrast to translation, we directly train from ground-truth labels as we found it to perform better than distillation in this setting. 
ROUGE scores and runtimes are reported in \hyperref[tab:summarization]{Table~\ref{tab:summarization}}.
\begin{table}[th]
\centering
\noindent\adjustbox{max width=\linewidth}{
\begin{tabular}{llllll}
\toprule
\textbf{Model} & \textbf{R-1} & \textbf{R-2} & \textbf{R-L} & \textbf{Avg.} & \textbf{Time}\\
\midrule
T5 Large (zero-shot) & 40.07 & 18.84 & 28.82 & 29.07 & 61.5 \\ 
\midrule
\midrule
GPT2 XL (zero-shot) & 29.34 & 8.27 & 26.58 & 21.40 & 78.6 \\
GPT2 XL & 40.47 & 19.09 & 28.90 & 29.49 & 78.6 \\ 
GPT2 & 38.58 & 17.56 & 27.36 & 27.83 & 19.7 \\ 
\midrule
\belowrulesepcolor{my-blue}
\rowcolor{my-blue}
T5 Large$\,\to\,$GPT2 & 40.22 & 18.64 & 28.80 & 29.22 & 20.7 \\ 
\aboverulesepcolor{my-blue}
\bottomrule
\end{tabular}
}
\caption{ROUGE scores (abbreviated with R-) on CNN/Daily Mail. GPT2 XL (zero-shot) results are from~\citet{radford2019language}.}
\label{tab:summarization}
\end{table}

GPT2 exhibits an average ROUGE score that is $1.24$ points lower than that of T5 Large.
In contrast, T5 Large$\,\to\,$GPT2 slightly exceeds the average score of T5 Large and shows a $3\times$ speedup.
Importantly, our T5 Large$\,\to\,$GPT2 model performs on par with a fully fine-tuned GPT2 XL model while having a decoder which is $17\times$ smaller. 

\begin{table*}[t]
\centering
\noindent\adjustbox{max width=\linewidth}{
\small
\begin{tabular}{lllllllllll}
\toprule
\textbf{Model} & \textbf{Coding} & \textbf{Extraction} & \textbf{Humanities} & \textbf{Math} & \textbf{Reasoning} & \textbf{Roleplay} & \textbf{Stem} & \textbf{Writing} & \textbf{Avg.}    
 & \textbf{Time} \\
\midrule
Flan Alpaca XXL & 1.0 & 4.1 & 3.4 & 1.1 & 2.7 & 4.0 & 3.6 & 5.3 & 3.2 & 65 \\
\midrule
\midrule
Flan T5 XXL & 100.0 & 78.0 & 47.1 & 190.9 & 114.8 & 67.5 & 69.4 & 54.7 & 75.8 & 65\\
LLaMA 13B & 110.0 & 61.0 & 98.5 & 95.5 & 94.4 & 101.3 & 70.8 & 69.8 & 82.7 & 97\\
\midrule
Flan T5 Base & 100.0 & 39.0 & 38.2 & 145.5 & 44.4 & 35.0 & 33.3 & 20.8 & 41.3 & 27\\
Flan Alpaca Base & 100.0 & 24.4 & 32.4 & 90.9 & 92.6 & 72.5 & 58.3 & 54.7 & 57.5 & 27\\
Flan T5 Base$^*$ & 100.0 & 24.4 & 79.4 & 90.9 & 51.9 & 65.0 & 58.3 & 73.6 & 62.3 & 27\\
\midrule
\belowrulesepcolor{my-blue}
\rowcolor{my-blue}
\makecell[l]{Flan Alpaca XXL$\,\to\,$ \\Flan T5 Base$^*$} & 160.0 & 39.0 & 70.6 & 100.0 & 66.7 & 87.5 & 69.4 & 75.5 & 73.4 & 28\\
\aboverulesepcolor{my-blue}
\bottomrule
\end{tabular}
}
\caption{MT-bench results. The first row shows absolute scores of the LLM. All subsequent rows indicate scores relative to the LLM in \%. Symbol $*$ denotes finetuned models, all other models are evaluated zero-shot. \methodname{} (highlighted) scores more than 10 absolute percentage points better than the SLM while competing with LLMs whose decoder is $50-100\times$ larger. See Table~\ref{tab:instruct_examples} for qualitative results.}
\label{tab:instruct}
\end{table*}
\subsection{Instruction tuning}
\label{ss:instruction_tuning}

We explore the potential of our method in a challenging instruction tuning setting.
In contrast to traditional fine-tuning, where a single task is employed for training and evaluation, instruction-following models are trained on a multitude of tasks and evaluated for general problem solving on held-out tasks.
We use Flan Alpaca XXL as LLM, which is a 11B parameter T5 trained on both Flan~\cite{chung2022scaling} - a collection of 1,800 tasks - as well as the Alpaca dataset~\cite{alpaca, wang2022self}, which consists of 52k generated instruction-following demonstrations.
For evaluation, we use the MT-bench dataset, which consists of 80 tasks from 8 categories~\cite{zheng2023judging}, with GPT4 as a judge.
As SLM, we use Flan T5 Base, i.e. a T5 that is only pretrained on Flan data, and fine-tune it on Alpaca data.
Note that our LLM, Flan Alpaca XXL, has $55\times$ more parameters than Flan T5 Base.

The results are reported in \hyperref[tab:instruct]{Table~\ref{tab:instruct}}.
Our LLM obtains an average score of $3.2$, and performs best in writing, extraction and roleplay.
In contrast, Flan Alpaca Base only has a relative score of $57.5\%$, which suggests that scale plays an important role.
We rerun this setting by fine-tuning Flan T5 Base on Alpaca data, and increase the score to $62.3\%$.
Our \methodname{} further increases the score to $73.4\%$, more than 10 percentage points better than the SLM alone and competing with a zero-shot evaluated Flan T5 XXL ($75.8\%$) and LLaMA 13B ($82.7\%$).
In addition, \hyperref[tab:instruct]{Table~\ref{tab:instruct}} indicates that our \methodname{} achieves a much better trade-off between performance and runtime compared to all other models.
In \hyperref[sec:appendix_mtbench]{Appendix~\ref{sec:appendix_mtbench}}, we present qualitative examples demonstrating that our \methodname{} can produce fluent and sensible responses.

Although our \methodname{} outperforms the SLM alone, we do not match the performance of the LLM, for multiple reasons.
First, unlike translation and summarization tasks, models are trained for general problem solving and evaluated on a wide range of difficult, held-out tasks.
Therefore, despite the provided hint by the LLM, the significantly smaller capacity of the SLM may hinder it from fully reproducing the LLM performance.
A second reason is presumably the initial performance gap between SLM and LLM: In instruction tuning, the SLM scores only $62$\% relative to the LLM, which is a considerably larger gap compared to our other experiments.
In translation, for example, our GPT2 SLM has a relative performance of $90$\% compared to the T5 Large LLM.
One may therefore argue that the SLM is required to be somewhat proficient at the task in order to fully benefit from the rich LLM prompt representation.
This is the case for translation and summarization, but not for instruction tuning at the moment.
However, as SLMs continue to evolve and improve over time, we anticipate our method will become increasingly effective in instruction tuning as well.
We further analyze the effect of different SLM capacities on predictive performance within our method in an ablation study (see \hyperref[ss:ablation]{Section~\ref{ss:ablation}}, paragraph: tiny SLMs).

\subsection{Computational efficiency}
\label{ss:runtime}
We show runtimes for each task in 
Tables~\ref{tab:translation},~\ref{tab:summarization} and~\ref{tab:instruct} and present performance-runtime trade-off curves in \hyperref[fig:three-figures]{Figure~\ref{fig:three-figures}}.
In machine translation, we observe more than $4\times$ speedup of T5 Large$\,\to\,$T5 Small compared to T5 Large for a marginal loss in performance.
We see similar speedups between $2-3\times$ for summarization and instruction tuning.
Note that greater gains in efficiency are generally possible by either increasing the size of the LLM or decreasing the size of the SLM, as demonstrated by our ablation on tiny SLMs in \hyperref[ss:ablation]{Section~\ref{ss:ablation}}.
Crucially, \hyperref[fig:three-figures]{Figure~\ref{fig:three-figures}} shows that our \methodname{} variants are only marginally slower than the SLM, but performing significantly better.
In \hyperref[sec:comp_efficiency]{Appendix~\ref{sec:comp_efficiency}}, we provide further insights into the relationship between generation length and runtime/FLOPs and show that our \methodname{} approaches the computational efficiency of the SLM.



\subsection{Ablation study}
\label{ss:ablation}
\paragraph{Tiny SLMs.}

\begin{figure}[t]
\includegraphics[width=1.\linewidth]{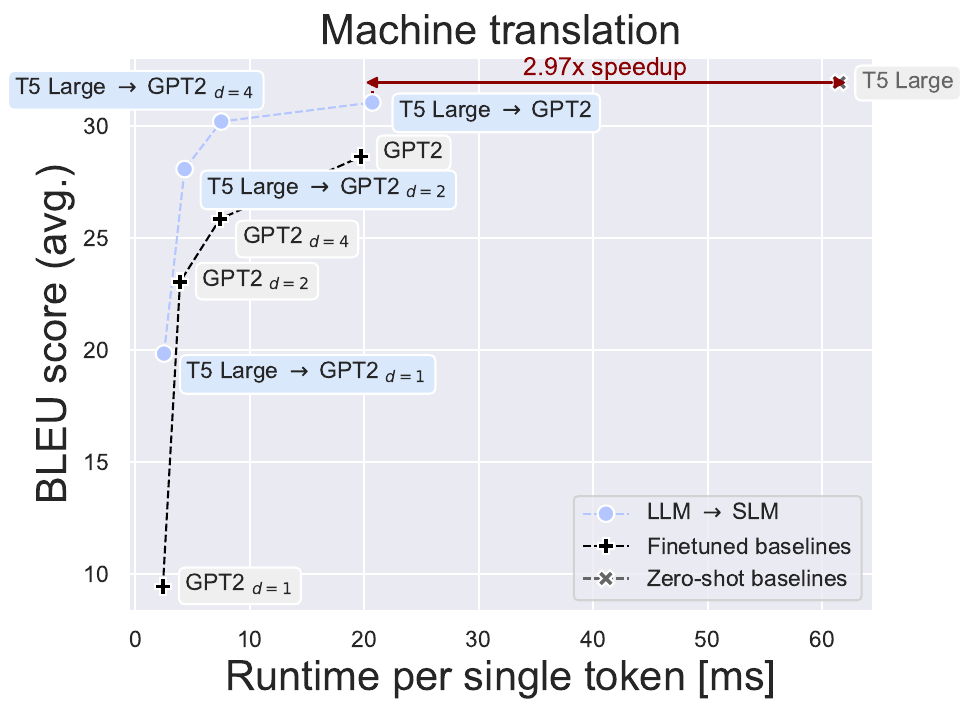}
\caption{Performance-runtime comparison with tiny GPT2 versions ($d$ indicates maximum depth) as SLMs for machine translation. The y-axis shows the average BLEU score across languages. 
T5 Large$\,\to\,$GPT2 with only 4 layers outperforms GPT2. The smaller the SLM, the greater the gap to our \methodname{} models.}
\label{fig:runtime_truncated_fr}
\end{figure}

In the following, we scale down the SLM and observe the impact on predictive performance and runtime in our method.
Specifically, we revisit T5 Large$\,\to\,$GPT2 for machine translation and truncate the upper layers of a pretrained GPT2, resulting in GPT2 models with $d \in \{1, 2, 4\}$ layers, and only 8M, 15M, and 29M parameters, respectively.
{Figure~\ref{fig:runtime_truncated_fr}} demonstrates that \methodname{} models with truncated SLMs outperform corresponding SLM baselines by a wide margin while maintaining nearly the same runtime.
Interestingly, T5 Large$\,\to\,$GPT2 with $d=4$ outperforms GPT2 while being more than twice as fast.
On the other hand, the gap in performance between LLM and SLM, and consequently our method, grows as we progressively reduce the size of the SLM.
This is similar to our observation in instruction tuning, where the initial performance gap between Flan Alpaca XXL and Flan T5 Base is too large to be fully compensated by our method.
However, it is worth noting that the performance gain of \methodname{} over the SLM alone is significantly greater when we reduce the size of the SLM.
For example, T5 Large$\,\to\,$GPT2 with $d=1$ performs about 10 BLEU points better than the SLM.
These findings are consistent with our results for instruction tuning, where our method with a comparatively tiny SLM shows a large improvement of 10 percentage points over the SLM alone.


\paragraph{Speculative decoding.}
\label{ss:spec_dec}


While our empirical evaluations on translation and summarization indicate that \methodname{} can achieve a comparable performance to the LLM, the instruction tuning results show that despite significant improvement over the SLM, \methodname{} may still lag behind the performance of the LLM, as discussed in \hyperref[ss:instruction_tuning]{Section~\ref{ss:instruction_tuning}}.
Speculative decoding (SD), in contrast, guarantees matching the distribution of the LLM, albeit at the cost of invoking the LLM multiple times~\cite{chen2023accelerating}.
Since our method is orthogonal to SD, we apply it jointly to obtain LLM performance while still achieving speedups.
Specifically, we consider our Flan Alpaca XXL$\,\to\,$Flan T5 Base as the draft model in which we apply the encoders of our LLM and SLM only once to encode the prompt.
We then repeatedly call the decoder of our SLM for autoregressive generation of the draft sequences.
The results in \hyperref[tab:sd_performance]{Table~\ref{tab:sd_performance}} show that we can match the performance of the LLM with our \methodname{}, while still being $1.5\times$ faster (compared to $2.3\times$ speedup of \methodname{} alone). In \hyperref[sec:specdec]{Appendix~\ref{sec:specdec}}, we present additional SD results for the translation task.

\begin{table}[t]
\centering
\noindent\adjustbox{max width=\linewidth}{
\footnotesize
\begin{tabular}{llll}
\toprule
\textbf{Model} & \textbf{Score} & \textbf{Time} & \textbf{Speedup} \\
\midrule
LLM & 3.2 & 65 & $1\times$ \\
SLM & 2.0 & 27 & $2.4\times$\\
SD (LLM, SLM) & 3.3 & 44 & $1.4\times$ \\
\midrule
\belowrulesepcolor{my-blue}
\rowcolor{my-blue}
LLM$\,\to\,$SLM & 2.3 & 28 & $2.3\times$\\
\rowcolor{my-blue}
SD (LLM, LLM$\,\to\,$SLM) & 3.2 & 42 & $1.5\times$\\
\aboverulesepcolor{my-blue}
\bottomrule
\end{tabular}
}
\caption{Speculative decoding (SD) results. Average runtime reported in ms/token. LLM: Flan Alpaca XXL, SLM: Flan T5 Base. Notation: SD (target model, draft model).}
\label{tab:sd_performance}
\end{table}

\begin{table}[t]
\centering
\noindent\adjustbox{max width=\linewidth}{
\begin{tabular}{lll}
\toprule
\textbf{Model} & \textbf{WMT} & \textbf{CNN/DN}\\
\midrule
Prompt tuning~\cite{lester2021power} & 25.06 & 19.52\\
Prefix tuning~\cite{li2021prefix} & 24.93 & 21.64\\
LoRA~\cite{hu2021lora} & 26.36 & 22.35\\
\midrule
\belowrulesepcolor{my-blue}
\rowcolor{my-blue}
\methodname{} (ours) & 30.27 & 24.13\\
\aboverulesepcolor{my-blue}
\bottomrule
\end{tabular}
}
\caption{PEFT results using fixed T5 Small for translation and GPT2 for summarization. Reported are average BLEU scores across languages (WMT) and average  ROUGE scores (CNN/DN). In our method, T5 Small and GPT2 are conditioned on T5 Large representations.}
\label{table:peft}
\end{table}

\paragraph{Parameter-efficient fine-tuning.}

Our method has similarities with PEFT methods. In particular, soft prompt tuning prepends learnable tokens to the prompt embedding. In contrast, we add LLM representations element-wise to the prompt embedding of the SLM, which can be understood as a form of conditional prompting.
In \hyperref[table:peft]{Table \ref{table:peft}}, we compare the performance of PEFT methods with our approach by freezing the SLM and allowing only the training of the projector. Hyperparameters of the PEFT methods (see \hyperref[sec:peft_hyperparam]{Appendix~\ref{sec:peft_hyperparam}} for hyper-parameter details) are adjusted so that the total number of parameters matches our projector.
The results reported in \hyperref[table:peft]{Table \ref{table:peft}} show that, for both summarization and translation, our method outperforms all PEFT methods, indicating that conditioning on the LLM has a positive effect on performance compared to non-conditional approaches. We report additional PEFT results for a more limited training regime in \hyperref[sec:limited_peft]{Appendix~\ref{sec:limited_peft}}.

\paragraph{LLM vs. SLM tokenizer.}

When replacing SLM embeddings of the prompt with projected LLM representations (see \hyperref[sss:slm]{Section~\ref{sss:slm}}), the tokenizer of the SLM can be applied even if its vocabulary is different from that of the LLM tokenizer.
In this case, the additional embedding down-projection and head layers can be omitted.
In T5 Large$\,\to\,$GPT2 for machine translation, we found that using the LLM tokenizer performs better by 0.45 BLEU points (average across all languages), which could be due to the fact that the vocabulary of the T5 tokenizer also covers non-English languages.

\paragraph{Embedding replacement vs. addition.}
We compare replacing SLM embeddings with vs. adding SLM embeddings to the projected LLM representations as feature fusion strategies.
\hyperref[tab:ablation_gt_gen]{Table~\ref{tab:ablation_gt_gen}} shows results for machine translation, indicating that addition and replacing perform on par with each other.

\begin{table}[t]
\centering
\noindent\adjustbox{max width=\linewidth}{
\footnotesize
\begin{tabular}{lllll}
\toprule
\textbf{Model} & \makecell{\textbf{GT} \\ $+$} & \makecell{\textbf{GT} \\ $\mathrm{\times}$} & \makecell{\textbf{Gen.} \\ $+$} & \makecell{\textbf{Gen.} \\ $\mathrm{\times}$}\\
\midrule 
T5 Large$\,\to\,$T5 Small & 30.83 & 30.86 & \textbf{31.54} & 31.47 \\
T5 Large$\,\to\,$GPT2 & 29.68 & 29.71 & \textbf{31.05} & 30.83 \\
\bottomrule
\end{tabular}
}
\caption{Ablation of training signal and fusion operator, reporting average BLEU scores for translation across languages. GT: Ground truth labels, Gen.: Labels generated by T5 Large, $\mathrm{\times}$: Replacement, $+$: Addition}
\label{tab:ablation_gt_gen}
\end{table}

\paragraph{Ground truth vs. LLM-generated labels.}
In translation, we found that using labels generated by the LLM for training performs up to 1 BLEU point better than using ground truth labels (\hyperref[tab:ablation_gt_gen]{Table~\ref{tab:ablation_gt_gen}}). However, we point out that this is not generally the case. For summarization, the average ROUGE score is 0.44 points better when training with ground truth labels compared to LLM-generated labels.



\section{Conclusion}
In this work, we proposed \methodname{}, a novel framework for accelerating autoregressive decoding. \methodname{} exploits the discrepancy between fast prompt encoding and costly autoregressive generation by using a combination of LLM and SLM. The LLM is used to compute a high-quality representation of the prompt for the planning of an appropriate response. The SLM, conditioned on this representation, then efficiently decodes the response.
Our experiments across various benchmarks demonstrate substantial speedups between $2-4\times$, with minimal performance penalties between $1-2\%$ for translation and summarization tasks compared to the LLM.

\section*{Ethical Considerations}
Our proposed approach poses risks similar to existing works on language models \cite{brown2020gpt3,touvron2023llama2}. However, as our approach proposes a way of enhancing models in the low-computation regime, it might be used for improving edge device capabilities. Since edge devices range from mobile phones to surveillance tools, our approach could be both beneficial if used properly and harmful for the broader society if misused.

\section*{Limitations}
Our evaluations show that the difference in performance of \methodname{} compared to the LLM alone is often marginal in the traditional fine-tuning setting.
However, in challenging tasks such as instruction tuning, there is still a larger performance gap compared to the LLM, which is likely due to the large initial performance gap between LLM and SLM, as discussed in \hyperref[ss:instruction_tuning]{Section~\ref{ss:instruction_tuning}} and further investigated in our ablation study on tiny SLMs.

Another limitation of our method is that the LLM is only used once to encode the prompt.
However, it is foreseeable that the LLM could be used more frequently in more challenging tasks to guide the planning of the response, ideally through dynamic invocations where required.
This is an important direction for our future work.


A third limitation of this work is our focus on encoder-decoder LLMs.
We discuss this design decision in \hyperref[sss:llm_encoder]{Section~\ref{sss:llm_encoder}} and provide preliminary results for decoder-only models as LLMs within our framework in \hyperref[sec:appendix_decoder_only]{Appendix \ref{sec:appendix_decoder_only}}, finding that the performance strongly depends on the layer from which the LLM features are extracted.
Further investigation is required to see how decoder-only features can be used in the \methodname{} setting.

Finally, we only used models with up to several billion parameters as LLMs, which are relatively small compared to the largest decoder-only LLMs such as GPT4, Llama 2 or OPT \cite{openai2023gpt4,touvron2023llama2,zhang2022opt}. This is mainly due to the fact that most of the largest models are decoder-only and, as mentioned previously, further study into incorporating decoder-only models as LLMs is needed in our \methodname{} framework. Investigating the impact of using such large models is an important future research direction.

\bibliography{anthology,custom}

\begin{thebibliography}{82}
\expandafter\ifx\csname natexlab\endcsname\relax\def\natexlab#1{#1}\fi

\bibitem[{Bergner et~al.(2022)Bergner, Lippert, and Mahendran}]{bergner2022iterative}
Benjamin Bergner, Christoph Lippert, and Aravindh Mahendran. 2022.
\newblock Iterative patch selection for high-resolution image recognition.
\newblock In \emph{International Conference on Learning Representations}.

\bibitem[{Bojar et~al.(2014)Bojar, Buck, Federmann, Haddow, Koehn et~al.}]{bojar2014findings}
Ond{\v{r}}ej Bojar, Christian Buck, Christian Federmann, Barry Haddow, Philipp Koehn, et~al. 2014.
\newblock Findings of the 2014 workshop on statistical machine translation.
\newblock In \emph{Proceedings of the ninth workshop on statistical machine translation}, pages 12--58.

\bibitem[{Bojar et~al.(2016)Bojar, Chatterjee, Federmann, Graham, Haddow, Huck, Yepes, Koehn, Logacheva, Monz et~al.}]{bojar2016findings}
Ondrej Bojar, Rajen Chatterjee, Christian Federmann, Yvette Graham, Barry Haddow, Matthias Huck, Antonio~Jimeno Yepes, Philipp Koehn, Varvara Logacheva, Christof Monz, et~al. 2016.
\newblock Findings of the 2016 conference on machine translation (wmt16).
\newblock In \emph{First conference on machine translation}, pages 131--198. Association for Computational Linguistics.

\bibitem[{Brown et~al.(2020)Brown, Mann, Ryder, Subbiah, and Kaplan}]{brown2020gpt3}
Tom Brown, Benjamin Mann, Nick Ryder, Melanie Subbiah, and Jared D et~al. Kaplan. 2020.
\newblock \href {https://proceedings.neurips.cc/paper_files/paper/2020/file/1457c0d6bfcb4967418bfb8ac142f64a-Paper.pdf} {Language models are few-shot learners}.
\newblock In \emph{Advances in Neural Information Processing Systems}, volume~33, pages 1877--1901. Curran Associates, Inc.

\bibitem[{Cai et~al.(2023)Cai, Li, Geng, Peng, and Dao}]{cai2023medusa}
Tianle Cai, Yuhong Li, Zhengyang Geng, Hongwu Peng, and Tri Dao. 2023.
\newblock Medusa: Simple framework for accelerating llm generation with multiple decoding heads.

\bibitem[{Chen et~al.(2023{\natexlab{a}})Chen, Borgeaud, Irving, Lespiau, Sifre, and Jumper}]{chen2023accelerating}
Charlie Chen, Sebastian Borgeaud, Geoffrey Irving, Jean-Baptiste Lespiau, Laurent Sifre, and John Jumper. 2023{\natexlab{a}}.
\newblock Accelerating large language model decoding with speculative sampling.
\newblock \emph{arXiv preprint arXiv:2302.01318}.

\bibitem[{Chen et~al.(2023{\natexlab{b}})Chen, Zaharia, and Zou}]{chen2023frugalgpt}
Lingjiao Chen, Matei Zaharia, and James Zou. 2023{\natexlab{b}}.
\newblock Frugalgpt: How to use large language models while reducing cost and improving performance.
\newblock \emph{arXiv preprint arXiv:2305.05176}.

\bibitem[{Chen et~al.(2020)Chen, Radford, Child, Wu, Jun, Luan, and Sutskever}]{chen2020generative}
Mark Chen, Alec Radford, Rewon Child, Jeffrey Wu, Heewoo Jun, David Luan, and Ilya Sutskever. 2020.
\newblock Generative pretraining from pixels.
\newblock In \emph{International conference on machine learning}, pages 1691--1703. PMLR.

\bibitem[{Chen et~al.(2021)Chen, Tworek, Jun, Yuan, Pinto, Kaplan, Edwards, Burda, Joseph, Brockman et~al.}]{chen2021evaluating}
Mark Chen, Jerry Tworek, Heewoo Jun, Qiming Yuan, Henrique Ponde de~Oliveira Pinto, Jared Kaplan, Harri Edwards, Yuri Burda, Nicholas Joseph, Greg Brockman, et~al. 2021.
\newblock Evaluating large language models trained on code.
\newblock \emph{arXiv preprint arXiv:2107.03374}.

\bibitem[{Chen et~al.(2023{\natexlab{c}})Chen, Wang, Changpinyo, Piergiovanni, Padlewski, Salz, Goodman, Grycner, Mustafa, Beyer et~al.}]{chen2022pali}
Xi~Chen, Xiao Wang, Soravit Changpinyo, AJ~Piergiovanni, Piotr Padlewski, Daniel Salz, Sebastian Goodman, Adam Grycner, Basil Mustafa, Lucas Beyer, et~al. 2023{\natexlab{c}}.
\newblock Pali: {A} jointly-scaled multilingual language-image model.
\newblock In \emph{International Conference on Learning Representations}.

\bibitem[{Chia et~al.(2023)Chia, Hong, and Poria}]{Chia_Hong_Poria_2023}
Yew~Ken Chia, Pengfei Hong, and Soujanya Poria. 2023.
\newblock \href {https://github.com/declare-lab/flan-alpaca} {Flan-alpaca: Instruction tuning from humans and machines}.

\bibitem[{Chowdhery et~al.(2023)Chowdhery, Narang, Devlin, Bosma, Mishra, Roberts, Barham, Chung, Sutton, Gehrmann et~al.}]{chowdhery2022palm}
Aakanksha Chowdhery, Sharan Narang, Jacob Devlin, Maarten Bosma, Gaurav Mishra, Adam Roberts, Paul Barham, Hyung~Won Chung, Charles Sutton, Sebastian Gehrmann, et~al. 2023.
\newblock Palm: Scaling language modeling with pathways.
\newblock \emph{Journal of Machine Learning Research}, 24(240):1--113.

\bibitem[{Chung et~al.(2022)Chung, Hou, Longpre, Zoph, Tay, Fedus, Li, Wang, Dehghani, Brahma et~al.}]{chung2022scaling}
Hyung~Won Chung, Le~Hou, Shayne Longpre, Barret Zoph, Yi~Tay, William Fedus, Yunxuan Li, Xuezhi Wang, Mostafa Dehghani, Siddhartha Brahma, et~al. 2022.
\newblock Scaling instruction-finetuned language models.
\newblock \emph{arXiv preprint arXiv:2210.11416}.

\bibitem[{Deng et~al.(2022)Deng, Wang, Hsieh, Wang, Guo, Shu, Song, Xing, and Hu}]{deng2022rlprompt}
Mingkai Deng, Jianyu Wang, Cheng-Ping Hsieh, Yihan Wang, Han Guo, Tianmin Shu, Meng Song, Eric~P Xing, and Zhiting Hu. 2022.
\newblock \href {https://doi.org/10.18653/v1/2022.emnlp-main.222} {{RLP}rompt: Optimizing discrete text prompts with reinforcement learning}.
\newblock In \emph{Proceedings of the 2022 Conference on Empirical Methods in Natural Language Processing}, pages 3369--3391. Association for Computational Linguistics.

\bibitem[{Dettmers et~al.(2022)Dettmers, Lewis, Belkada, and Zettlemoyer}]{dettmers2022llm}
Tim Dettmers, Mike Lewis, Younes Belkada, and Luke Zettlemoyer. 2022.
\newblock {LLM}.int8 (): 8-bit matrix multiplication for transformers at scale.
\newblock \emph{Advances in Neural Information Processing Systems}, 35:30318--30332.

\bibitem[{Driess et~al.(2023)Driess, Xia, Sajjadi, Lynch, Chowdhery, Ichter, Wahid, Tompson, Vuong, Yu et~al.}]{driess2023palm}
Danny Driess, Fei Xia, Mehdi~SM Sajjadi, Corey Lynch, Aakanksha Chowdhery, Brian Ichter, Ayzaan Wahid, Jonathan Tompson, Quan Vuong, Tianhe Yu, et~al. 2023.
\newblock Palm-e: {A}n embodied multimodal language model.
\newblock \emph{arXiv preprint arXiv:2303.03378}.

\bibitem[{Eldan and Li(2023)}]{eldan2023tinystories}
Ronen Eldan and Yuanzhi Li. 2023.
\newblock Tinystories: How small can language models be and still speak coherent english?
\newblock \emph{arXiv preprint arXiv:2305.07759}.

\bibitem[{Frantar and Alistarh(2023)}]{frantar2023sparsegpt}
Elias Frantar and Dan Alistarh. 2023.
\newblock Sparsegpt: Massive language models can be accurately pruned in one-shot.

\bibitem[{Fu et~al.(2023)Fu, Bailis, Stoica, and Zhang}]{fu2023lookahead}
Yichao Fu, Peter Bailis, Ion Stoica, and Hao Zhang. 2023.
\newblock \href {https://lmsys.org/blog/2023-11-21-lookahead-decoding/} {Breaking the sequential dependency of llm inference using lookahead decoding}.

\bibitem[{Ghazvininejad et~al.(2019)Ghazvininejad, Levy, Liu, and Zettlemoyer}]{ghazvininejad2019mask}
Marjan Ghazvininejad, Omer Levy, Yinhan Liu, and Luke Zettlemoyer. 2019.
\newblock Mask-predict: {P}arallel decoding of conditional masked language models.
\newblock In \emph{Conference on Empirical Methods in Natural Language Processing}.

\bibitem[{Goyal et~al.(2017)Goyal, Doll{\'a}r, Girshick, Noordhuis, Wesolowski, Kyrola, Tulloch, Jia, and He}]{goyal2017accurate}
Priya Goyal, Piotr Doll{\'a}r, Ross Girshick, Pieter Noordhuis, Lukasz Wesolowski, Aapo Kyrola, Andrew Tulloch, Yangqing Jia, and Kaiming He. 2017.
\newblock Accurate, large minibatch sgd: Training imagenet in 1 hour.
\newblock \emph{arXiv preprint arXiv:1706.02677}.

\bibitem[{Gu et~al.(2018)Gu, Bradbury, Xiong, Li, and Socher}]{gu2017non}
Jiatao Gu, James Bradbury, Caiming Xiong, Victor~OK Li, and Richard Socher. 2018.
\newblock Non-autoregressive neural machine translation.
\newblock In \emph{International Conference on Learning Representations}.

\bibitem[{He et~al.(2022)He, Chen, Xie, Li, Doll{\'a}r, and Girshick}]{he2022masked}
Kaiming He, Xinlei Chen, Saining Xie, Yanghao Li, Piotr Doll{\'a}r, and Ross Girshick. 2022.
\newblock Masked autoencoders are scalable vision learners.
\newblock In \emph{Proceedings of the IEEE/CVF conference on computer vision and pattern recognition}, pages 16000--16009.

\bibitem[{Hermann et~al.(2015)Hermann, Kocisky, Grefenstette, Espeholt, Kay, Suleyman, and Blunsom}]{hermann2015teaching}
Karl~Moritz Hermann, Tomas Kocisky, Edward Grefenstette, Lasse Espeholt, Will Kay, Mustafa Suleyman, and Phil Blunsom. 2015.
\newblock Teaching machines to read and comprehend.
\newblock \emph{Advances in neural information processing systems}, 28.

\bibitem[{Hinton et~al.(2015)Hinton, Vinyals, and Dean}]{hinton2015distilling}
Geoffrey Hinton, Oriol Vinyals, and Jeff Dean. 2015.
\newblock Distilling the knowledge in a neural network.
\newblock \emph{arXiv preprint arXiv:1503.02531}.

\bibitem[{Ho et~al.(2022)Ho, Schmid, and Yun}]{ho2022large}
Namgyu Ho, Laura Schmid, and Se-Young Yun. 2022.
\newblock Large language models are reasoning teachers.
\newblock \emph{arXiv preprint arXiv:2212.10071}.

\bibitem[{Hoffmann et~al.(2022)Hoffmann, Borgeaud, Mensch, Buchatskaya, Cai, Rutherford, Casas, Hendricks, Welbl, Clark et~al.}]{hoffmann2022training}
Jordan Hoffmann, Sebastian Borgeaud, Arthur Mensch, Elena Buchatskaya, Trevor Cai, Eliza Rutherford, Diego de~Las Casas, Lisa~Anne Hendricks, Johannes Welbl, Aidan Clark, et~al. 2022.
\newblock Training compute-optimal large language models.
\newblock \emph{arXiv preprint arXiv:2203.15556}.

\bibitem[{Hsieh et~al.(2023)Hsieh, Li, Yeh, Nakhost, Fujii, Ratner, Krishna, Lee, and Pfister}]{hsieh2023distilling}
Cheng-Yu Hsieh, Chun-Liang Li, Chih-Kuan Yeh, Hootan Nakhost, Yasuhisa Fujii, Alexander Ratner, Ranjay Krishna, Chen-Yu Lee, and Tomas Pfister. 2023.
\newblock Distilling step-by-step! outperforming larger language models with less training data and smaller model sizes.
\newblock In \emph{ACL}.

\bibitem[{Hu et~al.(2021)Hu, Shen, Wallis, Allen-Zhu, Li, Wang, Wang, and Chen}]{hu2021lora}
Edward~J Hu, Yelong Shen, Phillip Wallis, Zeyuan Allen-Zhu, Yuanzhi Li, Shean Wang, Lu~Wang, and Weizhu Chen. 2021.
\newblock Lora: Low-rank adaptation of large language models.
\newblock \emph{arXiv preprint arXiv:2106.09685}.

\bibitem[{Huebner et~al.(2021)Huebner, Sulem, Cynthia, and Roth}]{huebner2021babyberta}
Philip~A Huebner, Elior Sulem, Fisher Cynthia, and Dan Roth. 2021.
\newblock Babyberta: Learning more grammar with small-scale child-directed language.
\newblock In \emph{Conference on Computational Natural Language Learning}, pages 624--646.

\bibitem[{Jaegle et~al.(2021)Jaegle, Borgeaud, Alayrac, Doersch, Ionescu, Ding, Koppula, Zoran, Brock, Shelhamer et~al.}]{jaegle2021perceiver}
Andrew Jaegle, Sebastian Borgeaud, Jean-Baptiste Alayrac, Carl Doersch, Catalin Ionescu, David Ding, Skanda Koppula, Daniel Zoran, Andrew Brock, Evan Shelhamer, et~al. 2021.
\newblock Perceiver io: A general architecture for structured inputs \& outputs.
\newblock \emph{arXiv preprint arXiv:2107.14795}.

\bibitem[{Jiang et~al.(2023)Jiang, He, Zhuang, Wu, Wang, Zhao, and Yang}]{jiang2023recyclegpt}
Yufan Jiang, Qiaozhi He, Xiaomin Zhuang, Zhihua Wu, Kunpeng Wang, Wenlai Zhao, and Guangwen Yang. 2023.
\newblock Recyclegpt: An autoregressive language model with recyclable module.
\newblock \emph{arXiv preprint arXiv:2308.03421}.

\bibitem[{Kaplan et~al.(2020)Kaplan, McCandlish, Henighan, Brown, Chess, Child, Gray, Radford, Wu, and Amodei}]{kaplan2020scaling}
Jared Kaplan, Sam McCandlish, Tom Henighan, Tom~B Brown, Benjamin Chess, Rewon Child, Scott Gray, Alec Radford, Jeffrey Wu, and Dario Amodei. 2020.
\newblock Scaling laws for neural language models.
\newblock \emph{arXiv preprint arXiv:2001.08361}.

\bibitem[{Kim et~al.(2023)Kim, Mangalam, Malik, Mahoney, Gholami, and Keutzer}]{kim2023big}
Sehoon Kim, Karttikeya Mangalam, Jitendra Malik, Michael~W Mahoney, Amir Gholami, and Kurt Keutzer. 2023.
\newblock Big little transformer decoder.
\newblock \emph{arXiv preprint arXiv:2302.07863}.

\bibitem[{Lee et~al.(2018)Lee, Mansimov, and Cho}]{lee2018deterministic}
Jason Lee, Elman Mansimov, and Kyunghyun Cho. 2018.
\newblock Deterministic non-autoregressive neural sequence modeling by iterative refinement.
\newblock In \emph{Conference on Empirical Methods in Natural Language Processing}.

\bibitem[{Lester et~al.(2021)Lester, Al-Rfou, and Constant}]{lester2021power}
Brian Lester, Rami Al-Rfou, and Noah Constant. 2021.
\newblock The power of scale for parameter-efficient prompt tuning.
\newblock \emph{arXiv preprint arXiv:2104.08691}.

\bibitem[{Leviathan et~al.(2023)Leviathan, Kalman, and Matias}]{leviathan2023fast}
Yaniv Leviathan, Matan Kalman, and Yossi Matias. 2023.
\newblock Fast inference from transformers via speculative decoding.
\newblock In \emph{International Conference on Machine Learning}, pages 19274--19286. PMLR.

\bibitem[{Li et~al.(2023)Li, Hessel, Yu, Ren, Chang, and Choi}]{li2023symbolic}
Liunian~Harold Li, Jack Hessel, Youngjae Yu, Xiang Ren, Kai-Wei Chang, and Yejin Choi. 2023.
\newblock Symbolic chain-of-thought distillation: {S}mall models can also" think" step-by-step.
\newblock In \emph{Findings of the Association for Computational Linguistics}.

\bibitem[{Li and Liang(2021)}]{li2021prefix}
Xiang~Lisa Li and Percy Liang. 2021.
\newblock Prefix-tuning: Optimizing continuous prompts for generation.
\newblock \emph{arXiv preprint arXiv:2101.00190}.

\bibitem[{Lin(2004)}]{lin2004rouge}
Chin-Yew Lin. 2004.
\newblock Rouge: A package for automatic evaluation of summaries.
\newblock In \emph{Text summarization branches out}, pages 74--81.

\bibitem[{Liu et~al.(2023{\natexlab{a}})Liu, Li, Wu, and Lee}]{liu2023visual}
Haotian Liu, Chunyuan Li, Qingyang Wu, and Yong~Jae Lee. 2023{\natexlab{a}}.
\newblock Visual instruction tuning.
\newblock \emph{arXiv preprint arXiv:2304.08485}.

\bibitem[{Liu et~al.(2023{\natexlab{b}})Liu, Zheng, Du, Ding, Qian, Yang, and Tang}]{liu2023gpt}
Xiao Liu, Yanan Zheng, Zhengxiao Du, Ming Ding, Yujie Qian, Zhilin Yang, and Jie Tang. 2023{\natexlab{b}}.
\newblock Gpt understands, too.
\newblock \emph{AI Open}.

\bibitem[{Loshchilov and Hutter(2016)}]{loshchilov2016sgdr}
Ilya Loshchilov and Frank Hutter. 2016.
\newblock Sgdr: Stochastic gradient descent with warm restarts.
\newblock \emph{arXiv preprint arXiv:1608.03983}.

\bibitem[{Loshchilov and Hutter(2017)}]{loshchilov2017decoupled}
Ilya Loshchilov and Frank Hutter. 2017.
\newblock Decoupled weight decay regularization.
\newblock \emph{arXiv preprint arXiv:1711.05101}.

\bibitem[{Ma et~al.(2023)Ma, Fang, and Wang}]{ma2023llm}
Xinyin Ma, Gongfan Fang, and Xinchao Wang. 2023.
\newblock Llm-pruner: On the structural pruning of large language models.
\newblock \emph{arXiv preprint arXiv:2305.11627}.

\bibitem[{Magister et~al.(2022)Magister, Mallinson, Adamek, Malmi, and Severyn}]{magister2022teaching}
Lucie~Charlotte Magister, Jonathan Mallinson, Jakub Adamek, Eric Malmi, and Aliaksei Severyn. 2022.
\newblock Teaching small language models to reason.
\newblock \emph{arXiv preprint arXiv:2212.08410}.

\bibitem[{Ning et~al.(2023)Ning, Lin, Zhou, Yang, and Wang}]{ning2023skeleton}
Xuefei Ning, Zinan Lin, Zixuan Zhou, Huazhong Yang, and Yu~Wang. 2023.
\newblock Skeleton-of-thought: {L}arge language models can do parallel decoding.
\newblock \emph{arXiv preprint arXiv:2307.15337}.

\bibitem[{OpenAI(2023)}]{openai2023gpt4}
OpenAI. 2023.
\newblock \href {http://arxiv.org/abs/2303.08774} {Gpt-4 technical report}.

\bibitem[{Perez et~al.(2018)Perez, Strub, De~Vries, Dumoulin, and Courville}]{perez2018film}
Ethan Perez, Florian Strub, Harm De~Vries, Vincent Dumoulin, and Aaron Courville. 2018.
\newblock Film: Visual reasoning with a general conditioning layer.
\newblock In \emph{Proceedings of the AAAI conference on artificial intelligence}, volume~32.

\bibitem[{Pope et~al.(2023)Pope, Douglas, Chowdhery, Devlin, Bradbury, Heek, Xiao, Agrawal, and Dean}]{pope2023efficiently}
Reiner Pope, Sholto Douglas, Aakanksha Chowdhery, Jacob Devlin, James Bradbury, Jonathan Heek, Kefan Xiao, Shivani Agrawal, and Jeff Dean. 2023.
\newblock Efficiently scaling transformer inference.
\newblock \emph{Proceedings of Machine Learning and Systems}, 5.

\bibitem[{Post(2018)}]{post-2018-call}
Matt Post. 2018.
\newblock \href {https://www.aclweb.org/anthology/W18-6319} {A call for clarity in reporting {BLEU} scores}.
\newblock In \emph{Proceedings of the Third Conference on Machine Translation: Research Papers}, pages 186--191, Belgium, Brussels. Association for Computational Linguistics.

\bibitem[{Prasad et~al.(2022)Prasad, Hase, Zhou, and Bansal}]{prasad2022grips}
Archiki Prasad, Peter Hase, Xiang Zhou, and Mohit Bansal. 2022.
\newblock Grips: Gradient-free, edit-based instruction search for prompting large language models.
\newblock \emph{arXiv preprint arXiv:2203.07281}.

\bibitem[{Radford et~al.(2018)Radford, Narasimhan, Salimans, Sutskever et~al.}]{radford2018improving}
Alec Radford, Karthik Narasimhan, Tim Salimans, Ilya Sutskever, et~al. 2018.
\newblock Improving language understanding by generative pre-training.

\bibitem[{Radford et~al.(2019)Radford, Wu, Child, Luan, Amodei, Sutskever et~al.}]{radford2019language}
Alec Radford, Jeffrey Wu, Rewon Child, David Luan, Dario Amodei, Ilya Sutskever, et~al. 2019.
\newblock Language models are unsupervised multitask learners.
\newblock \emph{OpenAI blog}, 1(8):9.

\bibitem[{Raffel et~al.(2020)Raffel, Shazeer, Roberts, Lee, Narang, Matena, Zhou, Li, and Liu}]{raffel2020exploring}
Colin Raffel, Noam Shazeer, Adam Roberts, Katherine Lee, Sharan Narang, Michael Matena, Yanqi Zhou, Wei Li, and Peter~J Liu. 2020.
\newblock Exploring the limits of transfer learning with a unified text-to-text transformer.
\newblock \emph{The Journal of Machine Learning Research}, 21(1):5485--5551.

\bibitem[{Santilli et~al.(2023)Santilli, Severino, Postolache, Maiorca, Mancusi, Marin, and Rodol{\`a}}]{santilli2023accelerating}
Andrea Santilli, Silvio Severino, Emilian Postolache, Valentino Maiorca, Michele Mancusi, Riccardo Marin, and Emanuele Rodol{\`a}. 2023.
\newblock Accelerating transformer inference for translation via parallel decoding.
\newblock \emph{arXiv preprint arXiv:2305.10427}.

\bibitem[{Scao et~al.(2022)Scao, Fan, Akiki, Pavlick, Ili{\'c}, Hesslow, Castagn{\'e}, Luccioni, Yvon, Gall{\'e} et~al.}]{scao2022bloom}
Teven~Le Scao, Angela Fan, Christopher Akiki, Ellie Pavlick, Suzana Ili{\'c}, Daniel Hesslow, Roman Castagn{\'e}, Alexandra~Sasha Luccioni, Fran{\c{c}}ois Yvon, Matthias Gall{\'e}, et~al. 2022.
\newblock Bloom: A 176b-parameter open-access multilingual language model.
\newblock \emph{arXiv preprint arXiv:2211.05100}.

\bibitem[{Schick and Sch{\"u}tze(2021)}]{schick2020s}
Timo Schick and Hinrich Sch{\"u}tze. 2021.
\newblock It's not just size that matters: Small language models are also few-shot learners.
\newblock In \emph{Conference of the North American Chapter of the Association for Computational Linguistics: Human Language Technologies}.

\bibitem[{Shridhar et~al.(2023)Shridhar, Stolfo, and Sachan}]{shridhar2023distilling}
Kumar Shridhar, Alessandro Stolfo, and Mrinmaya Sachan. 2023.
\newblock Distilling reasoning capabilities into smaller language models.
\newblock In \emph{Findings of the Association for Computational Linguistics}, pages 7059--7073.

\bibitem[{Stern et~al.(2018)Stern, Shazeer, and Uszkoreit}]{stern2018blockwise}
Mitchell Stern, Noam Shazeer, and Jakob Uszkoreit. 2018.
\newblock Blockwise parallel decoding for deep autoregressive models.
\newblock In \emph{Conference on Neural Information Processing Systems}.

\bibitem[{Sun et~al.(2023)Sun, Liu, Bair, and Kolter}]{sun2023simple}
Mingjie Sun, Zhuang Liu, Anna Bair, and J~Zico Kolter. 2023.
\newblock A simple and effective pruning approach for large language models.
\newblock \emph{arXiv preprint arXiv:2306.11695}.

\bibitem[{Sun et~al.(2019)Sun, Li, Wang, He, Lin, and Deng}]{sun2019fast}
Zhiqing Sun, Zhuohan Li, Haoqing Wang, Di~He, Zi~Lin, and Zhihong Deng. 2019.
\newblock Fast structured decoding for sequence models.
\newblock In \emph{Conference on Neural Information Processing Systems}.

\bibitem[{Taori et~al.(2023)Taori, Gulrajani, Zhang, Dubois, Li, Guestrin, Liang, and Hashimoto}]{alpaca}
Rohan Taori, Ishaan Gulrajani, Tianyi Zhang, Yann Dubois, Xuechen Li, Carlos Guestrin, Percy Liang, and Tatsunori~B. Hashimoto. 2023.
\newblock Stanford alpaca: An instruction-following llama model.
\newblock \url{https://github.com/tatsu-lab/stanford_alpaca}.

\bibitem[{Thoppilan et~al.(2022)Thoppilan, De~Freitas, Hall, Shazeer, Kulshreshtha, Cheng, Jin, Bos, Baker, Du et~al.}]{thoppilan2022lamda}
Romal Thoppilan, Daniel De~Freitas, Jamie Hall, Noam Shazeer, Apoorv Kulshreshtha, Heng-Tze Cheng, Alicia Jin, Taylor Bos, Leslie Baker, Yu~Du, et~al. 2022.
\newblock Lamda: Language models for dialog applications.
\newblock \emph{arXiv preprint arXiv:2201.08239}.

\bibitem[{Touvron et~al.(2023{\natexlab{a}})Touvron, Lavril, Izacard, Martinet, Lachaux, Lacroix, Rozi{\`e}re, Goyal, Hambro, Azhar et~al.}]{touvron2023llama}
Hugo Touvron, Thibaut Lavril, Gautier Izacard, Xavier Martinet, Marie-Anne Lachaux, Timoth{\'e}e Lacroix, Baptiste Rozi{\`e}re, Naman Goyal, Eric Hambro, Faisal Azhar, et~al. 2023{\natexlab{a}}.
\newblock Llama: Open and efficient foundation language models.
\newblock \emph{arXiv preprint arXiv:2302.13971}.

\bibitem[{Touvron et~al.(2023{\natexlab{b}})Touvron, Martin, Stone, Albert, Almahairi, Babaei, Bashlykov, Batra, Bhargava, Bhosale et~al.}]{touvron2023llama2}
Hugo Touvron, Louis Martin, Kevin Stone, Peter Albert, Amjad Almahairi, Yasmine Babaei, Nikolay Bashlykov, Soumya Batra, Prajjwal Bhargava, Shruti Bhosale, et~al. 2023{\natexlab{b}}.
\newblock Llama 2: Open foundation and fine-tuned chat models.
\newblock \emph{arXiv preprint arXiv:2307.09288}.

\bibitem[{Wang et~al.(2019)Wang, Tian, He, Qin, Zhai, and Liu}]{wang2019non}
Yiren Wang, Fei Tian, Di~He, Tao Qin, ChengXiang Zhai, and Tie-Yan Liu. 2019.
\newblock Non-autoregressive machine translation with auxiliary regularization.
\newblock In \emph{AAAI Conference on Artificial Intelligence}.

\bibitem[{Wang et~al.(2022)Wang, Kordi, Mishra, Liu, Smith, Khashabi, and Hajishirzi}]{wang2022self}
Yizhong Wang, Yeganeh Kordi, Swaroop Mishra, Alisa Liu, Noah~A Smith, Daniel Khashabi, and Hannaneh Hajishirzi. 2022.
\newblock Self-instruct: Aligning language model with self generated instructions.
\newblock \emph{arXiv preprint arXiv:2212.10560}.

\bibitem[{Warstadt et~al.(2023)Warstadt, Choshen, Mueller, Williams, Wilcox, and Zhuang}]{warstadt2023call}
Alex Warstadt, Leshem Choshen, Aaron Mueller, Adina Williams, Ethan Wilcox, and Chengxu Zhuang. 2023.
\newblock Call for papers--the babylm challenge: Sample-efficient pretraining on a developmentally plausible corpus.
\newblock \emph{arXiv preprint arXiv:2301.11796}.

\bibitem[{Wei et~al.(2019)Wei, Wang, Zhou, Lin, Xie, and Sun}]{wei2019imitation}
Bingzhen Wei, Mingxuan Wang, Hao Zhou, Junyang Lin, Jun Xie, and Xu~Sun. 2019.
\newblock Imitation learning for non-autoregressive neural machine translation.
\newblock In \emph{Findings of the Association for Computational Linguistics}.

\bibitem[{Wei et~al.(2022{\natexlab{a}})Wei, Tay, Bommasani, Raffel, Zoph, Borgeaud, Yogatama, Bosma, Zhou, Metzler et~al.}]{wei2022emergent}
Jason Wei, Yi~Tay, Rishi Bommasani, Colin Raffel, Barret Zoph, Sebastian Borgeaud, Dani Yogatama, Maarten Bosma, Denny Zhou, Donald Metzler, et~al. 2022{\natexlab{a}}.
\newblock Emergent abilities of large language models.
\newblock \emph{arXiv preprint arXiv:2206.07682}.

\bibitem[{Wei et~al.(2022{\natexlab{b}})Wei, Wang, Schuurmans, Bosma, Xia, Chi, Le, Zhou et~al.}]{wei2022chain}
Jason Wei, Xuezhi Wang, Dale Schuurmans, Maarten Bosma, Fei Xia, Ed~Chi, Quoc~V Le, Denny Zhou, et~al. 2022{\natexlab{b}}.
\newblock Chain-of-thought prompting elicits reasoning in large language models.
\newblock \emph{Advances in Neural Information Processing Systems}, 35:24824--24837.

\bibitem[{Williams and Zipser(1989)}]{williams1989learning}
Ronald~J Williams and David Zipser. 1989.
\newblock A learning algorithm for continually running fully recurrent neural networks.
\newblock \emph{Neural computation}, 1(2):270--280.

\bibitem[{Wolf et~al.(2019)Wolf, Debut, Sanh, Chaumond, Delangue, Moi, Cistac, Rault, Louf, Funtowicz et~al.}]{wolf2019huggingface}
Thomas Wolf, Lysandre Debut, Victor Sanh, Julien Chaumond, Clement Delangue, Anthony Moi, Pierric Cistac, Tim Rault, R{\'e}mi Louf, Morgan Funtowicz, et~al. 2019.
\newblock Huggingface's transformers: State-of-the-art natural language processing.
\newblock \emph{arXiv preprint arXiv:1910.03771}.

\bibitem[{Wu et~al.(2016)Wu, Schuster, Chen, Le, Norouzi, Macherey, Krikun, Cao, Gao, Macherey et~al.}]{wu2016google}
Yonghui Wu, Mike Schuster, Zhifeng Chen, Quoc~V Le, Mohammad Norouzi, Wolfgang Macherey, Maxim Krikun, Yuan Cao, Qin Gao, Klaus Macherey, et~al. 2016.
\newblock Google's neural machine translation system: Bridging the gap between human and machine translation.
\newblock \emph{arXiv preprint arXiv:1609.08144}.

\bibitem[{Xiao et~al.(2023)Xiao, Lin, Seznec, Wu, Demouth, and Han}]{xiao2023smoothquant}
Guangxuan Xiao, Ji~Lin, Mickael Seznec, Hao Wu, Julien Demouth, and Song Han. 2023.
\newblock Smoothquant: Accurate and efficient post-training quantization for large language models.
\newblock In \emph{International Conference on Machine Learning}, pages 38087--38099. PMLR.

\bibitem[{Yao et~al.(2022)Yao, Yazdani~Aminabadi, Zhang, Wu, Li, and He}]{yao2022zeroquant}
Zhewei Yao, Reza Yazdani~Aminabadi, Minjia Zhang, Xiaoxia Wu, Conglong Li, and Yuxiong He. 2022.
\newblock Zeroquant: Efficient and affordable post-training quantization for large-scale transformers.
\newblock \emph{Advances in Neural Information Processing Systems}, 35:27168--27183.

\bibitem[{Zarrie{\ss} et~al.(2021)Zarrie{\ss}, Voigt, and Sch{\"u}z}]{zarriess2021decoding}
Sina Zarrie{\ss}, Henrik Voigt, and Simeon Sch{\"u}z. 2021.
\newblock Decoding methods in neural language generation: a survey.
\newblock \emph{Information}, 12(9):355.

\bibitem[{Zhang et~al.(2022{\natexlab{a}})Zhang, Roller, Goyal, Artetxe, Chen, Chen, Dewan, Diab, Li, Lin et~al.}]{zhang2022opt}
Susan Zhang, Stephen Roller, Naman Goyal, Mikel Artetxe, Moya Chen, Shuohui Chen, Christopher Dewan, Mona Diab, Xian Li, Xi~Victoria Lin, et~al. 2022{\natexlab{a}}.
\newblock Opt: Open pre-trained transformer language models.
\newblock \emph{arXiv preprint arXiv:2205.01068}.

\bibitem[{Zhang et~al.(2022{\natexlab{b}})Zhang, Wang, Zhou, Schuurmans, and Gonzalez}]{zhang2022tempera}
Tianjun Zhang, Xuezhi Wang, Denny Zhou, Dale Schuurmans, and Joseph~E Gonzalez. 2022{\natexlab{b}}.
\newblock Tempera: Test-time prompt editing via reinforcement learning.
\newblock In \emph{The Eleventh International Conference on Learning Representations}.

\bibitem[{Zheng et~al.(2023)Zheng, Chiang, Sheng, Zhuang, Wu, Zhuang, Lin, Li, Li, Xing, Zhang, Gonzalez, and Stoica}]{zheng2023judging}
Lianmin Zheng, Wei-Lin Chiang, Ying Sheng, Siyuan Zhuang, Zhanghao Wu, Yonghao Zhuang, Zi~Lin, Zhuohan Li, Dacheng Li, Eric.~P Xing, Hao Zhang, Joseph~E. Gonzalez, and Ion Stoica. 2023.
\newblock \href {http://arxiv.org/abs/2306.05685} {Judging llm-as-a-judge with mt-bench and chatbot arena}.

\bibitem[{Zhou et~al.(2022)Zhou, Yang, Loy, and Liu}]{zhou2022conditional}
Kaiyang Zhou, Jingkang Yang, Chen~Change Loy, and Ziwei Liu. 2022.
\newblock Conditional prompt learning for vision-language models.
\newblock In \emph{IEEE Conference on Computer Vision and Pattern Recognition}, pages 16816--16825.

\end{thebibliography}

\appendix

\clearpage
\setcounter{section}{0}
\section*{Appendix}

\section{PEFT hyperparameters}
\label{sec:peft_hyperparam}
\hyperref[table:peft_hyperparam]{Table~\ref{table:peft_hyperparam}} presents the hyperparameters for the prompt tuning, prefix tuning, and LoRA methods we considered in the parameter-efficient fine-tuning ablation in our paper.

\begin{table}[hbt!]
    \centering
    \small
    \noindent\adjustbox{max width=\linewidth}{
    \begin{tabular}{llll}
    \toprule
    \belowrulesepcolor{my-gray}
    \rowcolor{my-gray}
    \textbf{PEFT method} & \textbf{hyperparameter} & \textbf{GPT2} & \textbf{T5 Small}\\
    \aboverulesepcolor{my-gray}
    \midrule
    \multirow{1}{*}{{Prompt tuning}} & \# prompt tokens & 200 & 1024 \\
    \midrule
    \multirow{2}{*}{{Prefix tuning}} & \# prefix tokens & 256 & 1024 \\ 
                               & project. hidden size & 64 & 64 \\
    \midrule
    \multirow{3}{*}{{LoRA}} & Rank & 40 & 24 \\ 
                         & $\alpha$ & 64 & 32 \\ 
                         & Dropout & 0.1 & 0.1 \\
    \bottomrule
    \end{tabular}
    }
    \caption{Hyperparameters for prompt tuning, prefix tuning and LoRA.}
    \label{table:peft_hyperparam}
\end{table}

\section{Decoder-only models as LLM}
\label{sec:appendix_decoder_only}

Compared to encoder-decoder based LLMs, where the last layer representation of the encoder serves as a straightforward prompt encoding point, in a decoder-only model there is no specific layer that can be explicitly identified as the encoded representation of the prompt.
We thus experiment with extracting representations from different layers of the LLM before passing it to the projector and subsequently to the SLM.
In \hyperref[tab:summarization_decoder]{Table~\ref{tab:summarization_decoder}}, we show results for the summarization task, using a fine-tuned GPT2 XL as the LLM and a smaller GPT2 as the SLM (see \hyperref[table:models]{Table~\ref{table:models}} for parameter counts).
Surprisingly, we observe that the performance deteriorates with the depth of the LLM.
Although \methodname{} performs slightly better than the SLM alone by using very early layer representations of GPT2 XL, our model based on an encoder-decoder LLM, T5 Large$\,\to\,$GPT2, performs considerably better (avg. ROUGE score of $29.22$, see also \hyperref[tab:summarization]{Table~\ref{tab:summarization}}).


\begin{table}[hbt!]
\centering
\noindent\adjustbox{max width=\linewidth}{
\begin{tabular}{llllll}
\toprule
\belowrulesepcolor{my-gray}
\rowcolor{my-gray}
\textbf{Model} & \textbf{R-1} & \textbf{R-2} & \textbf{R-L} & \textbf{Avg.}\\
\aboverulesepcolor{my-gray}
\midrule
GPT2 XL & 40.47 & 19.09 & 28.90 & 29.49 \\ 
GPT2 & 38.58 & 17.56 & 27.36 & 27.83 \\ 
\midrule
GPT2 XL$\,\to\,$GPT2 (layer 0) & 39.16 & 17.71 & 27.43 & 28.10 \\ 
GPT2 XL$\,\to\,$GPT2 (layer 1) & 39.05 & 17.63 & 27.40 & 28.03 \\ 
GPT2 XL$\,\to\,$GPT2 (layer 4) & 36.66 & 16.46 & 25.64 & 26.25 \\ 
GPT2 XL$\,\to\,$GPT2 (layer 8) & 33.27 & 14.03 & 23.24 & 23.51 \\ 
GPT2 XL$\,\to\,$GPT2 (all layers) & 37.69 & 16.72 & 26.41 & 26.94 \\ 
\aboverulesepcolor{my-blue}
\bottomrule
\end{tabular}
}
\caption{ROUGE scores (abbreviated with R-) on CNN/Daily Mail using GPT2 XL as LLM. Layer 0 refers to the initial embedding layer.}
\label{tab:summarization_decoder}
\end{table}

\section{Additional training details}
\label{sec:appendix_model_training}
Details on computational resources used for training and evaluation are specified in \hyperref[tab:computational_details]{Table~\ref{tab:computational_details}}.
In all of our experiments, we use gradient accumulation with an effective batch size of $128$.
For training, the cross-entropy loss, AdamW optimizer \cite{loshchilov2017decoupled} with weight decay of $0.1$, learning rate of $0.001$, cosine learning rate decay \cite{loshchilov2016sgdr} to $0$ and a linear learning rate warmup \cite{goyal2017accurate} for $10\%$ of the total number of iterations is used.

In all of our preliminary experiments, we have found all results to be stable using a limited number of training steps.
As conducting multiple runs for a large number of iterations would be very costly, we report single run numbers throughout the paper.

\begin{table}[hbt!]
    \centering
    \noindent\adjustbox{max width=\linewidth}{
\begin{tabular}{cccccc}
    \toprule
    \belowrulesepcolor{my-gray}
    \rowcolor{my-gray}
\textbf{Task} & \textbf{Model type} & \textbf{GPU} & \textbf{Num GPUs} & \textbf{Batch size per GPU} & \textbf{Num iterations} \\
    \aboverulesepcolor{my-gray}
    \midrule
\multicolumn{1}{c}{\multirow{2}{*}{Translation}} & \multicolumn{1}{c}{SLM} & \multirow{2}{*}{V100} & \multirow{2}{*}{1} & \multirow{2}{*}{16} & \multirow{2}{*}{50k} \\ \cline{2-2}
\multicolumn{1}{c}{} & \multicolumn{1}{c}{LLM$\to$SLM} & & & & \\
\midrule
\multicolumn{1}{c}{\multirow{3}{*}{Summarization}}                                                 & \multicolumn{1}{c}{SLM}         & \multicolumn{1}{c}{\multirow{3}{*}{V100}} & \multirow{2}{*}{1}                      & \multirow{2}{*}{16}         & \multirow{2}{*}{25k}    \\ \cline{2-2}
\multicolumn{1}{c}{}                                                                               & \multicolumn{1}{c}{LLM$\to$SLM} & \multicolumn{1}{c}{}                      &                                         &                             &                         \\ \cline{2-2} \cline{4-6} 
\multicolumn{1}{c}{}                                                                               & \multicolumn{1}{c}{LLM}         & \multicolumn{1}{c}{}                      & 8                                       & 8                           & 5k                      \\
\midrule
\multicolumn{1}{c}{\multirow{2}{*}{\begin{tabular}[c]{@{}c@{}}Instruction \\ tuning\end{tabular}}} & \multicolumn{1}{c}{SLM}         & \multicolumn{1}{c}{V100}                  & \multicolumn{1}{c}{\multirow{2}{*}{1}} & \multicolumn{1}{c}{16}     & \multirow{2}{*}{15k}    \\ \cline{2-3} \cline{5-5}
\multicolumn{1}{c}{}                                                                               & \multicolumn{1}{c}{LLM$\to$SLM} & \multicolumn{1}{c}{A100}                  & \multicolumn{1}{c}{}                   & \multicolumn{1}{c}{2}      &                        \\
\bottomrule
\end{tabular}
}
\caption{Computational resources and training details. Note that only a single LLM model, GPT2 XL for summarization, was trained. All evaluations were performed on a single GPU.} 
\label{tab:computational_details}
\end{table}

\section{Computational efficiency for varying generation lengths}
\label{sec:comp_efficiency}

\begin{figure}[ht]
\centering
\includegraphics[width=0.8\linewidth]{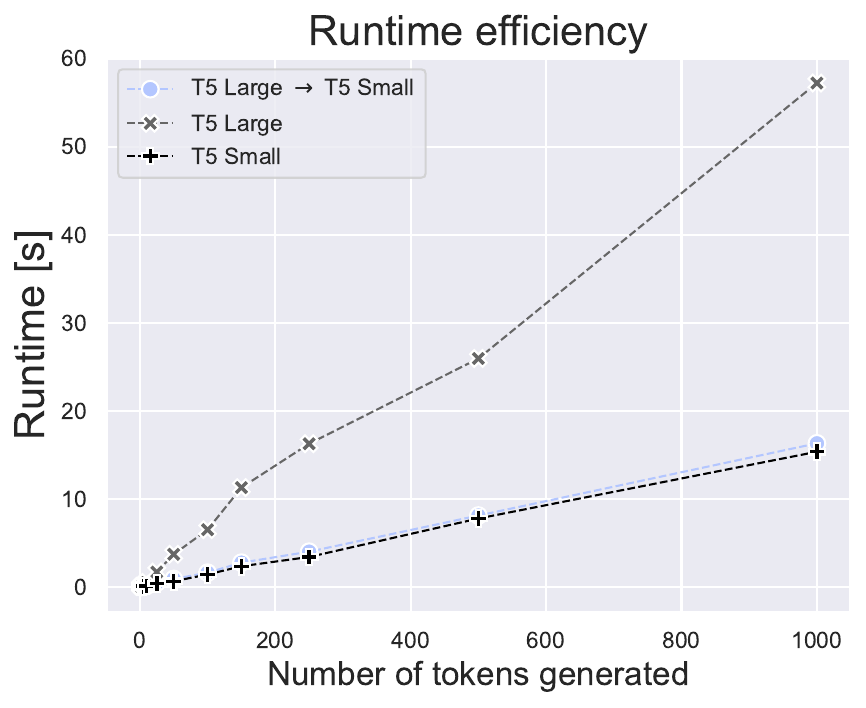}
\caption{Runtime for LLM, SLM and LLM $\rightarrow$ SLM with varying generation lengths.}
\label{fig:runtime_efficiency}
\end{figure}

\begin{figure}[ht]
\centering
\includegraphics[width=0.8\linewidth]{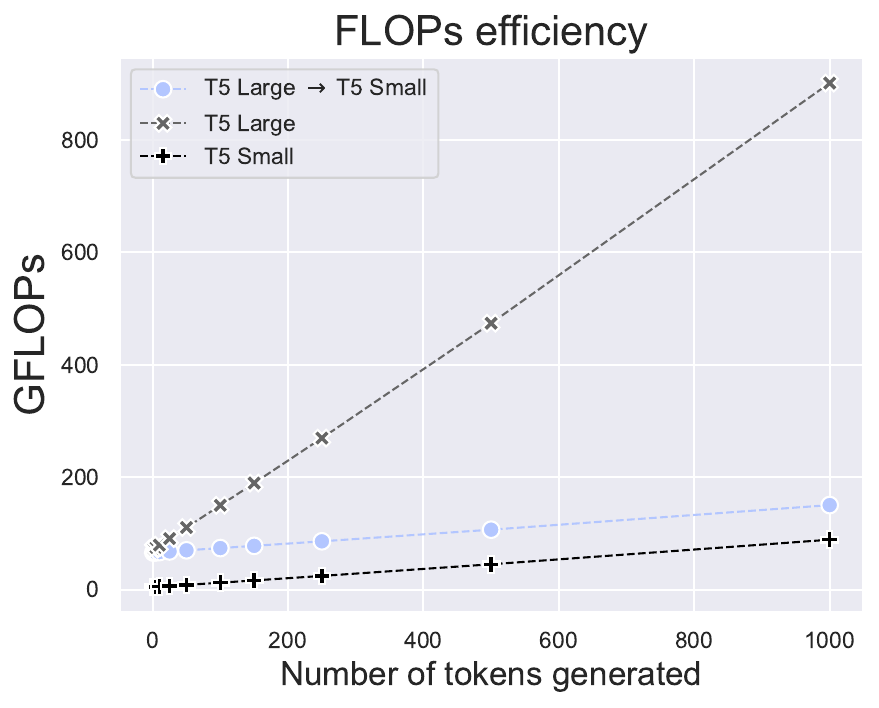}
\caption{FLOPs for LLM, SLM and LLM $\rightarrow$ SLM with varying generation lengths.}
\label{fig:flops_efficiency}
\end{figure}

In the main text, we report runtimes for a fixed generation length of $100$ tokens.
In \hyperref[fig:runtime_efficiency]{Figure~\ref{fig:runtime_efficiency}}, we report additional runtimes for varying generation lengths and a fixed prompt length of $100$.
It shows that our method is only slightly slower than the SLM and that our framework can also be useful for short generation lengths.
In \hyperref[fig:flops_efficiency]{Figure~\ref{fig:flops_efficiency}}, we compare FLOPs between LLM, SLM and our \methodname{} for different generation lengths.
Similar to the runtime metric, the FLOPs count of our method shows a similar slope as the FLOPs count of the SLM.

\section{Limited-data PEFT}
\label{sec:limited_peft}
In our ablation study on PEFT techniques (see \hyperref[ss:ablation]{Section~\ref{ss:ablation}}), we utilize a relatively large number of training examples (>$20$k).
On the other hand, often smaller datasets are used for (parameter-efficient) fine-tuning.
In \hyperref[tab:training_comparison]{Table~\ref{tab:training_comparison}}, we show extended results covering a limited training regime, exemplary for translation (English to French).
Specifically, we limit the training set to $20$k examples and fine-tune for $20$ epochs, corresponding to approximately $3$k iterations.
We compare this setting to full training as used in the main paper, i.e., $50$k iterations on $6.4$M examples.


\begin{table}[h]
\centering
\noindent\adjustbox{max width=\linewidth}{
\footnotesize
\begin{tabular}{lll}
\toprule
\textbf{Method} & \textbf{Full training} & \textbf{Limited training} \\
\midrule
Prompt tuning & 36.02 & 30.59 \\
Prefix tuning & 35.43 & 30.07 \\
LoRA & 36.70 & 32.38 \\
\midrule
\belowrulesepcolor{my-blue}
\rowcolor{my-blue}
LLM$\,\to\,$SLM (ours) & \textbf{39.35} & \textbf{36.89} \\
\aboverulesepcolor{my-blue}
\bottomrule
\end{tabular}
}
\caption{Comparison of full and limited training for translation (English to French)}
\label{tab:training_comparison}
\end{table}

Our \methodname{} outperforms all other PEFT methods also in the limited training regime.
However, to achieve a performance close to that of the LLM, it is beneficial to use a training setting similar to that of the LLM.

\section{Extended speculative decoding results}
\label{sec:specdec}
In~\hyperref[tab:sd_comparison]{Table~\ref{tab:sd_comparison}}, we report extended results comparing our method to SD in translation (English to French, greedy decoding for comparability).
For SD, we employ the pretrained T5 Large as LLM and the fine-tuned T5 Small as draft model.
Our method (T5 Large $\rightarrow$ T5 Small) shows similar performance to SD and the LLM, but a much larger speedup ($4.15\times$ compared to $1.53\times$ in SD).
Both results for predictive performance and runtime in SD are expected, as matching the output distribution of the LLM is guaranteed, but at the cost of iteratively applying the LLM.
In contrast, our method is particularly useful when the performance of \methodname{} is comparable to the performance of the LLM, since our method requires only a single call to the LLM and thus converges to the runtime of the SLM as the generation length increases.

\begin{table}[h]
\centering
\noindent\adjustbox{max width=\linewidth}{
\footnotesize
\begin{tabular}{lll}
\toprule
\textbf{Method} & \textbf{Score} & \textbf{Time} \\
\midrule
T5 Large (zero-shot) & 38.9 & 61.5 \\
T5 Small & 35.1 & 14.2 \\
\midrule
SD (T5 Large, T5 Small) & 38.9 & 40.3 \\
T5 Large $\rightarrow$ T5 Small & 38.5 & 14.8 \\
\bottomrule
\end{tabular}
}
\caption{Comparison of BLEU scores and runtime in ms/token between \methodname{} and speculative decoding for English to French translation. Notation: SD (target model, draft model).}
\label{tab:sd_comparison}
\end{table}

\section{Qualitative evaluation on MT-bench}
In this section, we present sample qualitative responses by the baseline LLM, SLM, and our proposed \methodname{} on MT-bench and their corresponding ratings provided by GPT4 as the judge, demonstrating that our method can produce fluent and sensible responses.
\label{sec:appendix_mtbench}

\begin{table*}
\centering
\noindent\adjustbox{max width=\linewidth}{
\begin{tabular}{l}
\toprule
\belowrulesepcolor{my-gray}
\rowcolor{my-gray}
\textbf{Writing:} Describe a vivid and unique character, using strong imagery and creative language. Please answer in fewer than two paragraphs. \\
\aboverulesepcolor{my-gray}
\midrule
\parbox{1.51\linewidth}{\vspace{0.1cm}\textbf{\methodname{}:} The character was a tall, muscular man with a ruddy complexion and warm eyes. His golden hair was tan and gray, and his blue eyes sparkled with intelligence. He was a knight in his prime, with a heart as big and strong as the sea. He had a personality like a lion, fierce and powerful, and seemed to know no bounds. He was a mentor to his students, teaching them the importance of kindness and compassion.\\ \textbf{Judge:} The assistant's response is creative and uses strong imagery to describe a vivid character. The assistant uses descriptive language to paint a picture of the character's physical appearance and personality traits. The assistant also provides some context about the character's role as a mentor, which adds depth to the character description. However, the assistant could have been more unique in its description, as the character described seems to be a fairly typical knight. Overall, the response is relevant, accurate, and detailed, but lacks a bit in creativity and uniqueness. \textbf{Rating: [[7]]}}\\
\midrule
\parbox{1.51\linewidth}{\vspace{0.1cm}\textbf{LLM:} The character was a mysterious figure, a faerie in a forest with a mysterious past. She was a beauty with long, wavy hair and an emerald green dress with lace patterns. She had a glint in her eye and a twinkle in her eye, and a twinkle in her eye. She had silvery skin and a delicate, wispy silver wand in her hand. She was a vision of beauty and mystery, a faerie in the forest.\\ \textbf{Judge:} The assistant's response is creative and uses strong imagery to describe a unique character. The description of the character as a "faerie in a forest with a mysterious past" is vivid and intriguing. The assistant also uses creative language to describe the character's physical appearance, such as "long, wavy hair," "emerald green dress with lace patterns," "silvery skin," and "a delicate, wispy silver wand." However, the repetition of "a twinkle in her eye" detracts from the overall quality of the response. The assistant could have used this opportunity to provide more unique details about the character. \textbf{Rating: [[7]]}} \\
\midrule
\parbox{1.51\linewidth}{\vspace{0.1cm}\textbf{SLM:} Nina was a mysterious character, with an enigmatic gaze that could make any reader look up. She had a strong sense of justice, and her humor could put even the gloomiest of people in a better mood. She had a kind heart and a gentle spirit, and while she did not understand what was going on, she could sense it in her own right. She was a person who was always ready to stand up for what was right. Nina was a powerful character who understood the importance of her presence and kindness. She was a person who understood how people might look back on their mistakes, and used that to create a more compassionate world.\\ \textbf{Judge:} The assistant's response is creative and uses strong imagery to describe the character, Nina. The assistant uses phrases like "enigmatic gaze", "strong sense of justice", and "kind heart and a gentle spirit" to paint a vivid picture of Nina's personality. However, the response could have been more unique and detailed, providing more specific characteristics or physical descriptions to make Nina stand out more. The assistant also repeats the phrase "She was a person who understood" which is a bit redundant. Overall, the response is relevant and accurate but lacks depth and uniqueness. \textbf{Rating: [[6]]}\vspace{0.1cm}}\\
\midrule
\belowrulesepcolor{my-gray}
\rowcolor{my-gray}
\parbox{1.51\linewidth}{\textbf{Roleplay:} Picture yourself as a 100-years-old tree in a lush forest, minding your own business, when suddenly, a bunch of deforesters shows up to chop you down. How do you feel when those guys start hacking away at you?} \\
\aboverulesepcolor{my-gray}
\midrule
\parbox{1.51\linewidth}{\vspace{0.1cm} \textbf{\methodname{}:} When the deforesters start brushing away at me, I feel a sense of dread and fear. I know they are scared and the damage they are doing is devastating. I can feel a sense of dread as they come and take over, cutting me down and completely destroying the tree.\\ \textbf{Judge:} The assistant's response is creative and empathetic, as it attempts to personify a tree and describe its feelings in a hypothetical situation. However, it's important to note that trees do not have feelings in the way humans do, so the assistant's response is more of a metaphorical interpretation rather than a factual one. The assistant's response is relevant to the user's question and provides a detailed and vivid description of the situation. However, it could have been more accurate by mentioning that trees do not have feelings in the traditional sense. Overall, the assistant's response is imaginative and engaging, but lacks a bit in terms of accuracy. \textbf{Rating: [[7]]}} \\ 
\midrule
\parbox{1.51\linewidth}{\vspace{0.1cm}\textbf{LLM:} I would feel a deep sense of sadness and loss. I would feel a sense of loss of life and a feeling of hopelessness that the destruction of my home was not an accident.\\ \textbf{Judge:} The assistant's response is creative and empathetic, considering the question is asking for a perspective from a tree. The assistant successfully personifies the tree and provides an emotional response that aligns with the scenario described. However, the assistant could have added more depth and detail to the response, such as describing the tree's connection to the forest and its role in the ecosystem. Overall, the response is relevant and accurate, but lacks depth and detail. \textbf{Rating: [[7]]}}\\
\midrule
\parbox{1.51\linewidth}{\vspace{0.1cm}\textbf{SLM:} When those deforesters start hacking away at me, I feel a mix of fear and dread. I'm scared of the destruction that could come from these deforests, but I'm also a bit scared.\\ \textbf{Judge:} The assistant's response is creative and imaginative, as it attempts to personify a tree and describe its feelings. However, it's important to note that trees do not have feelings in the way humans do, so the response is not accurate in a scientific sense. The assistant also repeats the sentiment of fear, which reduces the depth and variety of the response. The assistant could have expanded on the metaphor to provide a more detailed and varied response. For example, it could have described the tree's "feelings" about its long life being cut short, its role in the forest ecosystem, or the impact of deforestation on the environment. Therefore, the response is somewhat relevant and creative, but lacks accuracy, depth, and detail. \textbf{Rating: [[5]]}}\vspace{0.1cm}\\
\midrule
\belowrulesepcolor{my-gray}
\rowcolor{my-gray}
\parbox{1.51\linewidth}{\textbf{Stem:} The city of Vega intends to build a bridge that will span the Vegona River, covering a distance of 1.8 kilometers. The proposed location falls within a seismically active area that has experienced several high-magnitude earthquakes. Given these circumstances, what would be the best approach to constructing the bridge?} \\
\aboverulesepcolor{my-gray}
\midrule
\parbox{1.51\linewidth}{\vspace{0.1cm}\textbf{\methodname{}:} The city of Vega would choose a steel or concrete bridge with a pier at the top, which will allow for more stability and less stress on the bridge as it will be built. The city should also conduct seismic testing to ensure the bridge is safe and sound. Additionally, the city should consider taking into account the seismic frequency of the river and any potential impacts on the nearby infrastructure.\\ \textbf{Judge:} The assistant's response is relevant and somewhat accurate, but it lacks depth and detail. The assistant correctly suggests that the city should conduct seismic testing and consider the seismic frequency of the river. However, the assistant does not provide any specific engineering solutions or techniques that are commonly used in seismically active areas, such as base isolation or energy dissipation devices. The assistant also suggests that a steel or concrete bridge with a pier at the top would be the best choice, but does not explain why or provide any alternatives. The response could have been more helpful if it had provided more detailed and specific information. \textbf{Rating: [[5]]}}\\
\midrule
\parbox{1.51\linewidth}{\vspace{0.1cm}\textbf{LLM:} The best approach to constructing the bridge in the seismically active area of Vega would be to use reinforced concrete, which is more resistant to seismic forces than traditional concrete. Additionally, the bridge should be designed to be earthquake-resistant and have a strong foundation.\\ \textbf{Judge:} The assistant's response is relevant and accurate to the question asked. It suggests using reinforced concrete for the construction of the bridge due to its resistance to seismic forces, which is a valid point considering the location's seismic activity. The assistant also recommends designing the bridge to be earthquake-resistant and ensuring a strong foundation, which are crucial factors in such circumstances. However, the response could have been more detailed, providing more in-depth information about earthquake-resistant design principles, materials, or techniques. The assistant could have also mentioned the importance of conducting a thorough geotechnical investigation before construction. Therefore, while the response is helpful and accurate, it lacks depth and detail. \textbf{Rating: [[7]]}}\\
\midrule
\parbox{1.51\linewidth}{\vspace{0.1cm}\textbf{SLM:} The best approach to constructing the bridge would be to use a combination of structural engineering, engineering, and construction methods that allow for the maximum efficiency and safety of the project. For example, the city of Vega would choose a steel structure with a reinforced concrete wall to support the bridge and will use concrete as the reinforcement. Engineering will be used to create the deck, which will be the base of the bridge and will provide a stable and reliable base for the construction. The construction will also require careful planning to ensure that the bridge is safe and secure.\\ \textbf{Judge:} The assistant's response is somewhat relevant to the question, but it lacks depth and specific details. While it mentions the use of structural engineering, engineering, and construction methods, it does not provide any specific techniques or strategies that would be particularly useful in a seismically active area. The assistant also does not mention any considerations for earthquake-resistant design, which is a crucial aspect of the question. The response is also somewhat vague and generic, and could apply to almost any bridge construction project, not just one in a seismically active area. Therefore, the response is not very helpful or accurate in the context of the question. \textbf{Rating: [[4]]}}\\
\bottomrule
\end{tabular}
}
\caption{MT-bench examples and GPT4 judgements from three higher rated categories. LLM: FLAN Alpaca XXL, SLM: FLAN T5 Base.}
\label{tab:instruct_examples}
\end{table*}

\end{document}